%% file: IEEE-conference-template-062824.tex
\documentclass[conference]{IEEEtran}
\IEEEoverridecommandlockouts

\usepackage{cite}
\usepackage{amsmath,amssymb,amsfonts}
\usepackage{graphicx}
\usepackage{textcomp}
\usepackage{xcolor}
\usepackage[table]{xcolor}

\usepackage{booktabs}
\usepackage{multirow}
\usepackage{colortbl}
\usepackage{xcolor}

\usepackage{listings}

\usepackage{algorithm}
\usepackage{algpseudocode}
\usepackage[table]{xcolor}
\algnewcommand\algorithmicinput{\textbf{Input:}}
\algnewcommand\Input{\item[\algorithmicinput]}

\algnewcommand\algorithmicoutput{\textbf{Output:}}
\algnewcommand\Output{\item[\algorithmicoutput]}

\lstdefinestyle{paperStyle}{
    language=Python,
    basicstyle=\ttfamily\small,       
    keywordstyle=\bfseries,           
    commentstyle=\itshape\color{gray},
    stringstyle=\color{black},        
    frame=single,                     
    rulecolor=\color{black},          
    framerule=0.8pt,                  
    aboveskip=5pt,                    
    belowskip=15pt,                   
    showstringspaces=false,           
    breaklines=true,                  
    numbers=none                      
}

\def\BibTeX{{\rm B\kern-.05em{\sc i\kern-.025em b}\kern-.08em
    T\kern-.1667em\lower.7ex\hbox{E}\kern-.125emX}}

\usepackage{tikz}
\usepackage{pgfplots}
\usepackage{algorithm}
\usepackage{algpseudocode}
\usepackage[skins]{tcolorbox}
\definecolor{geoPrimary}{RGB}{0, 128, 128}   
\definecolor{geoDark}{RGB}{0, 60, 60}        
\definecolor{geoPurple}{RGB}{112, 48, 160}   

\definecolor{geoVermillion}{RGB}{213, 94, 0} 
\definecolor{geoOchre}{RGB}{230, 159, 0}     

\definecolor{geoSky}{RGB}{70, 150, 200}      
\definecolor{geoSage}{RGB}{0, 158, 115}      

\definecolor{geoCharcoal}{RGB}{50, 50, 50}   
\definecolor{geoSilver}{RGB}{180, 180, 180}  
\definecolor{geoFog}{RGB}{240, 242, 245}     

\newtcolorbox{takeawaybox}[2][]{
  enhanced,
  boxrule=0pt, frame hidden,      
  borderline west={4pt}{0pt}{geoPrimary}, 
  colback=geoPrimary!7!white,     
  coltitle=geoDark,               
  fonttitle=\bfseries\small,      
  title={#2},                     
  attach title to upper={\par\vspace{2pt}}, 
  sharp corners,                  
  #1
}

\usepackage{tikz}
\usepackage{tikz-3dplot}
\usepackage{amsmath, amssymb}
\usetikzlibrary{shapes.geometric, arrows.meta, positioning, calc, shadows.blur, fit, backgrounds, patterns, decorations.markings, bending}

\definecolor{geoPrimary}{RGB}{0, 128, 128}    
\definecolor{geoDark}{RGB}{0, 60, 60}           
\definecolor{geoLight}{RGB}{230, 245, 245}    
\definecolor{geoAccent}{RGB}{34, 139, 34}      
\definecolor{linearColor}{RGB}{70, 110, 180}  
\definecolor{paperWhite}{RGB}{250, 250, 250}  

\tikzset{
    dataNode/.style={
        rectangle, draw=gray!40, fill=white, line width=0.8pt,
        minimum width=0.7cm, minimum height=0.35cm,
        inner sep=1pt,
        font=\tiny\sffamily
    },
    layerDense/.style={
        rectangle, rounded corners=2pt, draw=geoDark!80, fill=geoPrimary!20,
        minimum width=0.8cm, minimum height=2.2cm, line width=0.8pt,
        drop shadow={opacity=0.15}, align=center
    },
    layerInternal/.style={
        rectangle, rounded corners=2pt, draw=geoDark!80, fill=geoPrimary!10,
        minimum width=0.6cm, minimum height=1.4cm, line width=0.5pt, align=center
    },
    layerResContainer/.style={
        rectangle, rounded corners=4pt, draw=geoDark, fill=white,
        line width=1pt,
        drop shadow={opacity=0.15, shadow xshift=2pt, shadow yshift=-2pt}
    },
    opCircle/.style={
        circle, draw=geoDark, fill=white, thick, inner sep=1pt,
        minimum size=0.5cm, font=\bfseries\small,
        drop shadow={opacity=0.1}
    },
    normBlock/.style={
        trapezium, trapezium angle=80, draw=geoAccent, fill=white,
        line width=1.2pt, minimum width=2.4cm, minimum height=0.9cm,
        align=center, font=\bfseries\scriptsize, text=geoDark,
        drop shadow={opacity=0.15}
    },
    flow/.style={->, >=latex, thick, draw=geoDark!80, rounded corners=4pt},
    residFlow/.style={->, >=latex, thick, dashed, draw=geoDark!60, rounded corners=4pt},
    vec/.style={->, >=latex, thick, line cap=round},
    clientBox/.style={
        rectangle, draw=geoDark, fill=white, thick,
        minimum width=2.4cm, minimum height=1.0cm, 
        align=center, font=\sffamily\bfseries\scriptsize,
        drop shadow={opacity=0.1}
    },
    serverBox/.style={
        rectangle, rounded corners=4pt, draw=linearColor, fill=linearColor!10, thick,
        minimum width=3.6cm, minimum height=2.8cm, 
        align=center, font=\sffamily\bfseries\small, text=linearColor!80!black,
        drop shadow={opacity=0.15}
    },
    squircleStrip/.style={
        rectangle, rounded corners=18pt, line width=1.2pt,
        inner sep=0pt,
        shadow xshift=2pt, shadow yshift=-2pt,
        drop shadow={opacity=0.04}
    },
    stripLabel/.style={
        font=\bfseries\sffamily, rotate=90, anchor=south, 
        inner sep=10pt, scale=0.9
    }
}

\usepgfplotslibrary{groupplots}
\pgfplotsset{compat=1.18}
\usepackage{eso-pic}
   \usepackage[hidelinks]{hyperref}
\def\IEEEDOI{10.1109/IJCNN69854.2026.11728825}
\AddToShipoutPictureBG*{%
  \AtPageLowerLeft{%
    \put(\LenToUnit{0.6in},\LenToUnit{0.3in}){%
      \parbox{\dimexpr\paperwidth-1.2in\relax}{\centering\scriptsize
        \copyright~2026 IEEE. Article published in the 2026 International Joint Conference on Neural Networks (IJCNN 2026).\\The record is available at \href{https://doi.org/\IEEEDOI}{https://doi.org/\IEEEDOI}.}}}}
\begin{document}
\title{Geodesic-SVDD Latent Mixup for One-Shot Distributed Tabular Learning\thanks{Code available at: \url{https://github.com/andreamoleri/Geodesic-SVDD}}
}

\author{
    \IEEEauthorblockN{
        Andrea Moleri\textsuperscript{1,2,*}, 
        Christian Internò\textsuperscript{3,*}, 
        Markus Olhofer\textsuperscript{1}, 
        Barbara Hammer\textsuperscript{3},
        and Ali Raza\textsuperscript{1} 
    }
    \IEEEauthorblockA{
        \textsuperscript{1}Honda Research Institute Europe, \space
        \textsuperscript{2}University of Milan-Bicocca, \space
        \textsuperscript{3}Bielefeld University\\
        \textsuperscript{*}These authors contributed equally to this work.
    }
}
\maketitle

\begin{abstract}
Distributed Learning on structured data is a key and widely adopted approach in privacy-sensitive sectors, ranging from financial fraud prevention to federated healthcare diagnostics. 
While Gradient Boosted Decision Trees are widely used in centralized settings due to their efficiency, they are incompatible with standard gradient-based federated learning protocols because of their non-differentiability. Conversely, standard deep classifiers that are amenable to federated optimization often suffer from catastrophic forgetting under label distribution skew.
In this work, to address these gaps we propose Geodesic-SVDD, a framework for Distributed Spherical Prototype Learning that recasts each client's task as one-class density estimation on the hypersphere, so that local models can be aggregated as probabilistic manifolds in a single round.
Specifically, for each client, we employ a Residual Spherical Encoder and Geodesic Latent Mixup that enforces strict hyperspherical constraints ($\|\phi(x)\| \equiv 1$), preventing the manifold collapse observed in linear approximations.
Furthermore, leveraging this insight, we propose a novel One-Shot ``Product of Experts'' protocol that enables privacy-preserving aggregation without iterative synchronization.
Finally, we mechanistically analyze the trade-off between interpolation strategies, demonstrating that while linear mixup offers an enticing computational shortcut, it destabilizes optimization gradients by orders of magnitude ($10^9\times$).
Across five tabular benchmarks and four levels of label skew, Geodesic-SVDD attains the highest mean AUROC among seven federated methods, outperforms federated tree ensembles by up to 21 AUROC points on \textit{Covertype}, and its One-Shot variant matches centralized Deep SVDD under IID and mild skew with a single communication round.
\end{abstract}


\section{Introduction}

Distributed Learning, often referred to as Federated Learning (FL)~\cite{mcmahan2017communication} on complex structured data (e.g., tabular) is a key requirement for safety and privacy sensitive sectors, ranging from classification tasks in financial fraud prevention to federated healthcare diagnostics. In centralized environments, Gradient Boosted Decision Trees (GBDTs) and Isolation Forests are the \textit{de facto} standard due to their robustness on irregular manifolds~\cite{grinsztajn2022tree}. However, deploying these models in distributed classification presents what we term the ``\textit{Federated Tabular Gap}.'' Tree ensembles are inherently non-differentiable and difficult to aggregate privacy-preservingly without sharing raw data splits or generating synthetic proxies~\cite{li2020federated}. 
This limits their accuracy on complex benchmarks such as \textit{Covertype}, as our experiments show (Sec.~\ref{sec:experiments}).
Deep Learning offers a differentiable alternative via weights aggregation algorithms like \texttt{FedAvg}~\cite{mcmahan2017communication}. However, standard deep classifiers are notoriously brittle under \textit{Label Distribution Skew}, where clients possess disjoint subsets of the global classes~\cite{kairouz2021advances}. Averaging the weights of neural networks trained on disjoint manifolds results in destructive interference and leads to poor global generalization performance~\cite{kirkpatrick2017overcoming}, which renders the global model unreliable for critical decision-making. Furthermore, these methods typically assume fully labeled datasets, which are rarely available in real edge deployment.

To bridge this gap, we propose \textit{Geodesic-SVDD}, a framework for Distributed Spherical Prototype Learning. Unlike standard classifiers that partition space with hyperplanes, our approach treats local learning as an \textit{unsupervised density estimation problem}~\cite{ruff2018deep}. We map inputs to a compact prototype on a hyperspherical manifold ($\mathbb{S}^{d-1}$), allowing clients to learn valid feature representations without requiring contrasting class labels. We then aggregate these local experts via a \textit{``Product of Experts''} (PoE) protocol~\cite{hinton2002training}.

Crucially, this architecture eases the traditional trade-off between communication efficiency~\cite{interno2024adaptive} and model robustness. 
We demonstrate that by aggregating unsupervised probabilistic manifolds rather than supervised weights, we can approach the performance of centralized baselines using a single-round One-Shot protocol. 
This capability significantly reduces the communication burden in non-IID settings, limiting the exposure to iterative gradient leakage~\cite{zhu2019deep} while matching a model trained on pooled data under IID and mild skew.
A critical design choice for algorithmic stability is the enforcement of geometric integrity during local training. 
To handle data sparsity, we employ \textit{Geodesic Latent Mixup}. 
While our empirical analysis reveals that standard Linear Interpolation~\cite{zhang2017mixup} can achieve competitive predictive scores, we demonstrate that such approximations induce \textit{Manifold Collapse} (midpoint norm $0.71$ for random pairs) and severe gradient instability ($10^9\times$ gradient variance).
By strictly interpolating along the Riemannian manifold via Spherical Linear Interpolation (SLERP)~\cite{shoemake1985animating}, Geodesic-SVDD ensures that high detection performance is grounded in valid boundary definitions, a prerequisite for reliable deployment.

The main contributions of this paper are:
\begin{itemize}
    \item \textbf{Unsupervised One-Shot Learning:} 
    We propose a novel architecture that aggregates unsupervised probabilistic manifolds through a Product of Experts. With a single communication round, it matches centralized Deep SVDD under IID and mild skew (mean AUROC within $0.6$ points) on tabular benchmarks.
    \item \textbf{Non-IID Robustness:}
    Under label skewness, Geodesic-SVDD (Iter) stays within $1.5$ AUROC points of its IID performance on every benchmark, attains the highest mean AUROC among federated methods, and outperforms federated tree ensembles by up to 21 AUROC points on \textit{Covertype}.
\item \textbf{Geometric Analysis:} We demonstrate that linear approximations destabilize optimization (gradient variance $3.99\times10^{-3}$ vs. $5.5\times10^{-12}$) and induce manifold collapse (midpoint norm $0.71$ vs. $1.00$ for random pairs), whereas the proposed Geodesic approach enforces strict constraints to ensure gradient stability and valid boundary definitions.
\end{itemize}

\section{Background, Key Challenges and Objectives}
\label{sec:challenges}
We identify three challenges to deploying robust, privacy-preserving models in non-IID distributed tabular environments:

\subsubsection{\textbf{The Tabular Task}}
In centralized settings, Gradient Boosted Decision Trees (GBDTs) dominate tabular learning due to their robustness on irregular manifolds~\cite{grinsztajn2022tree, borisov2021deep}. However, their non-differentiability creates a ``\textit{Federated Tabular Gap},'' obstructing gradient aggregation without costly cryptography or synthetic proxies~\cite{cheng2021secureboost}. As shown in Sec.~\ref{sec:experiments}, federated tree ensembles plateau at $0.71$--$0.76$ AUROC on \textit{Covertype}, where our method reaches $0.91$. A differentiable framework recovering tree-like robustness is therefore required.

\subsubsection{\textbf{Non-IID Scenarios}}
Real-world federated data frequently exhibits \textit{Label Distribution Skew}. In extreme cases, clients observe disjoint class subsets, causing standard \texttt{FedAvg} to suffer from weight divergence that grows with the degree of skew~\cite{zhao2018federated, 10651455}. To address this, we recast local learning as single-prototype density estimation, in the spirit of \textit{prototype aggregation}~\cite{tan2022fedproto}.

\subsubsection{\textbf{The Constraints of Iterative Communication}}
Standard iterative FL requires persistent bi-directional channels, widening the attack surface for deep leakage reconstruction~\cite{zhu2019deep}.  While ``One-Shot'' protocols minimize this exposure~\cite{guha2019one}, aggregating disjoint experts without iterative alignment remains a significant open challenge.


\subsection{The Classical Federated Setting}
We consider a distributed classification task with $K$ heterogeneous clients holding private datasets $\mathcal{D}_k = \{(x_i, y_i)\}_{i=1}^{N_k}$. In safety-critical scenarios, data is often mutually exclusive ($\mathcal{Y}_i \cap \mathcal{Y}_j = \emptyset$). Under this constraint, simple averaging fails as clients solve fundamentally different problems~\cite{zhao2018federated}.
Standard FL seeks a global weight vector $\omega^*$ by minimizing:
\begin{equation}
    \min_{\omega} \mathcal{L}_{Fed}(\omega) = \sum_{k=1}^K \frac{N_k}{N} \mathbb{E}_{(x,y) \sim \mathcal{D}_k} [\ell_{ce}(\mathcal{F}(x; \omega), y)]
\end{equation}
Under disjoint support, this objective is ill-posed; optimizing $\omega_k$ for $\mathcal{Y}_k$ effectively maximizes entropy for unseen classes.

\subsection{Optimization Objective}
\label{sec:objective}
To resolve the conflict between label skew and weight averaging, we decompose the multiclass problem into $C$ distinct \textit{One-vs-All} density estimation tasks. 

For a target class $c_{tgt} \in \mathcal{Y}$, we define a strictly \textit{Single-Center} objective. 
Let $\mathcal{D}_k^{+} = \{x \mid (x,y) \in \mathcal{D}_k, y = c_{tgt}\}$ denote the local target subset. Crucially, labels are used strictly for partitioning, keeping local learning as unsupervised density estimation~\cite{ruff2018deep}.

We formulate the global inference as an \textit{Anomaly Score} $\mathcal{A}(x)$ derived from a \textit{Product of Experts (PoE)} aggregation:
\begin{align}
    \label{eq:global_obj}
    \mathcal{A}(x; c_{tgt}) &= - \sum_{k=1}^K \log p(x \mid \theta_k, \mathbf{c}) \\
    \label{eq:local_opt}
    \text{where } \theta_k &= \operatorname*{argmin}_{\theta} \mathbb{E}_{x \sim \mathcal{D}_k^{+}} \left[ \| \phi(x; \theta) - \mathbf{c} \|^2 \right]
\end{align}
subject to $\|\phi(x)\|_2 \equiv 1$. The density $p(x)$ is modeled via a Student-t kernel ($\nu=1$) centered at $\mathbf{c}$.
This satisfies two requirements: \textbf{Single-Center Compactness} (minimizing distance to $\mathbf{c}$ without class-conditional prototypes) and \textbf{Decoupled Aggregation} (summing negative log-densities to form the global score).

\section{Methodology}
\label{sec:methodology}
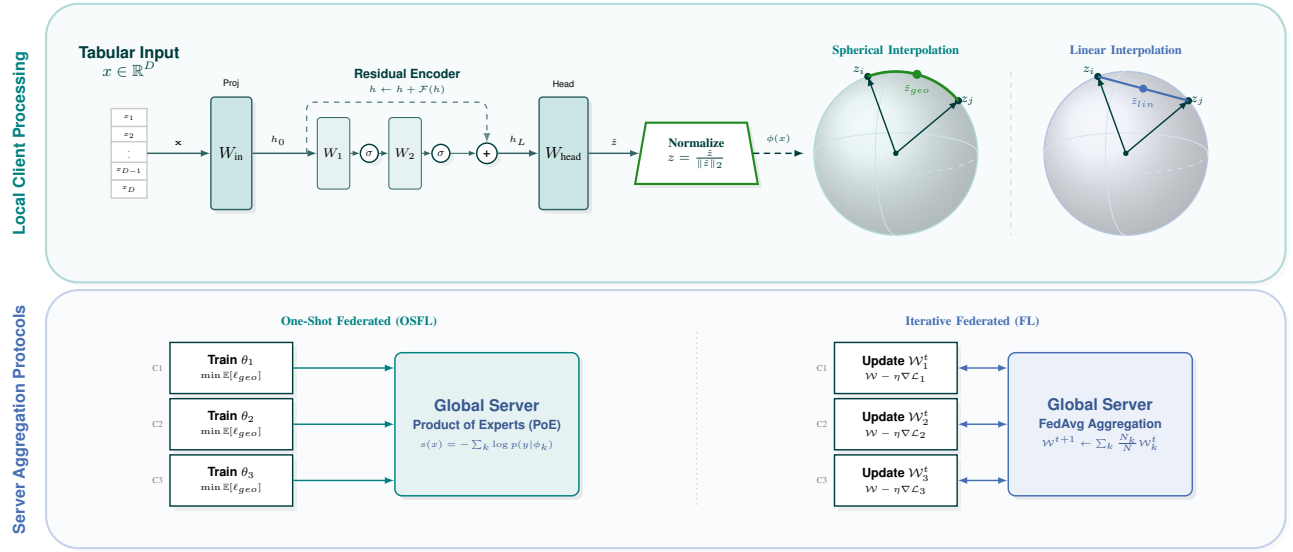
\begin{figure*}[t] 
    \centering
    \resizebox{0.95\textwidth}{!}{%
        \input{diagram.tex}
    }
    \caption{\textbf{The Geodesic-SVDD Framework.} \textbf{(Top)} Local clients utilize a Residual Spherical Encoder to map sparse tabular features onto the unit hypersphere $\mathbb{S}^{d-1}$. We contrast our Geodesic Mixup (which maintains unit norm via SLERP) against Linear Mixup (which collapses the manifold norm). \textbf{(Bottom)} The architecture supports two protocols: privacy-preserving One-Shot aggregation via Product of Experts (PoE) and iterative Federated Averaging.}
    \label{fig:federated_arch}

\end{figure*}

As illustrated in Fig.~\ref{fig:federated_arch}, our framework consists of two distinct operational stages: \textit{Local Client Processing}, where disjoint tabular partitions are mapped to a hyperspherical manifold, and \textit{Server Aggregation}, which supports both One-Shot (PoE) and Iterative protocols.

\subsection{Local Client Processing}
\label{sec:local_processing}
Each client $k$ possesses a local dataset $\mathcal{D}_k$. We strictly define the local objective as unsupervised Single-Center SVDD targeting a unique fixed center $c$; labels are utilized solely for partition support to define $\mathcal{D}_k$, ensuring that the global ensemble (Eq. \ref{eq:global_obj}) aggregates compatible density estimates rather than discriminative boundaries. The goal is to map discrete, sparse tabular features $x \in \mathbb{R}^D$ into a compact latent representation $\phi(x)$ on the unit hypersphere $\mathbb{S}^{d-1}$. This formulation recasts the objective from discriminative boundary learning to \textit{one-class density estimation}

\subsubsection{\textbf{Residual Spherical Encoder}}
Standard deep networks map data to an unconstrained Euclidean space $\mathbb{R}^d$. To enforce geometric rigor, we employ a dedicated encoder architecture (Fig.~\ref{fig:federated_arch}, Top-Left).
First, a linear projection $W_{in}$ expands the input to the hidden dimension $h_0$. This is passed through a sequence of $L$ Residual Blocks. The update rule for the $l$-th block is defined as $h_{l+1} = h_l + \mathcal{F}(h_l; W_1, W_2)$, where $\mathcal{F}$ represents the non-linear residual path containing LayerNorm, GELU activations ($\sigma$), and dropout.
Finally, the representation is projected by a head $W_{head}$ to a pre-normalized vector $\hat{z}$. 

Crucially, we enforce the manifold constraint via an explicit normalization layer:
\begin{equation}
    z = \phi(x) = \frac{\hat{z}}{\| \hat{z} \|_2}, \quad \text{where } \hat{z} = W_{head} \cdot h_L
\end{equation}
This ensures that all valid latent points lie strictly on the surface of the sphere ($\|z\| \equiv 1$), converting the optimization problem to a geodesic distance minimization.

\subsubsection{\textbf{Geodesic vs. Linear Interpolation}}
To robustly define decision boundaries on this manifold under data sparsity, we employ latent mixup. As depicted in Fig.~\ref{fig:federated_arch} (Top-Right), standard linear interpolation (LERP~\cite{zhang2017mixup, verma2019manifold}) takes the chordal path $\tilde{z}_{lin} = \lambda z_i + (1-\lambda)z_j$.
This operation violates the manifold geometry, as $\|\tilde{z}_{lin}\| < 1$ for any $\lambda \in (0,1)$, collapsing the synthetic sample into the interior of the sphere.

To preserve geometric integrity, we employ \textit{Spherical Linear Interpolation} (SLERP~\cite{shoemake1985animating}), which traverses the geodesic arc:
\begin{equation}
    \tilde{z}_{geo} = \frac{\sin((1-\lambda)\Omega)}{\sin(\Omega)}z_i + \frac{\sin(\lambda\Omega)}{\sin(\Omega)}z_j
\end{equation}
where $\Omega = \arccos(z_i \cdot z_j)$. This ensures $\|\tilde{z}_{geo}\| \equiv 1$, maintaining the encoder's output constraint.

\subsection{Server Aggregation Protocols}
The framework supports two aggregation strategies (Fig.~\ref{fig:federated_arch}, Bottom Panel) tailored to different safety and bandwidth constraints.

\subsubsection{\textbf{Protocol A: One-Shot Federated (OSFL)}}
Clients independently train their local models $\theta_k$ to minimize the geodesic error. To ensure global alignment without iterative synchronization, the optimization target $c$ is pre-computed via a warm-up phase. This requires a one-time exchange of latent feature statistics ($\mathbb{E}[\phi(x)]$) based on the shared random initialization, ensuring the global center is centrally aligned before local training begins. Specifically, clients compute local feature means using the shared random initialization $\theta_{init}$. These are aggregated at the server to form the fixed center $c \leftarrow \frac{\bar{\mu}}{\|\bar{\mu}\|}$, where $\bar{\mu} = \frac{1}{K} \sum_{k=1}^K \mathbb{E}_{x \sim \mathcal{D}_k^{+}}[\phi(x; \theta_{init})]$. 
The server then aggregates the trained ``experts'' into a global density estimator using a Product of Experts (PoE) rule $s(x) = - \sum_{k=1}^K \log p(x \mid \theta_k)$. This sums the log-likelihoods (densities) of the disjoint experts, forming a union of manifolds without requiring weight alignment.

\subsubsection{\textbf{Protocol B: Iterative Federated (FL)}}
For scenarios permitting bi-directional communication, we employ standard weight averaging. Crucially, strictly adhering to the One-vs-All protocol, the single target prototype $c$ is ``fixed'' ($c \leftarrow \frac{\bar{\mu}}{\|\bar{\mu}\|}$) on the hypersphere prior to training, as explicitly defined in Algorithm~\ref{alg:unified_svdd}.
This ensures that all clients map their disjoint data partitions to a consistent global coordinate system for the target class, preventing the rotation misalignment common in standard FL.
In round $t$, the server averages client weights $W^t_{k}$ to produce a global model $W^{t+1}$. This aligns the local coordinate systems but increases the gradient attack surface.

\begin{algorithm}[t]
\caption{\small Geodesic-SVDD: Unified Training Protocol}
\label{alg:unified_svdd}
\small
\begin{algorithmic}[1]
    \Input Datasets $\{\mathcal{D}_k\}_{k=1}^K$, Target Class $y_{tgt}$
    \Input \textbf{Config:} $\mathcal{G} \in \{\textsc{Slerp}, \textsc{Lerp}\}$, $\mathcal{P} \in \{\textsc{OneShot}, \textsc{Iterative}\}$
    \Output Anomaly Decision $\hat{y} \in \{0, 1\}$ for Target Class $y_{tgt}$

    \State \textbf{Filter Support:} $\mathcal{D}_k^+ \leftarrow \{x \mid (x, y) \in \mathcal{D}_k, y = y_{tgt}\}$
    \State \textbf{Server:} Init $\theta_g$; \textbf{Clients:} Compute global center $\mathbf{c}$
    \State \textbf{Warm-up:}
    \State \quad $\mathbf{c} \leftarrow \frac{\bar{\mu}}{\|\bar{\mu}\|}$ where $\bar{\mu} = \frac{1}{K} \sum_{k=1}^K \mathbb{E}_{x \sim \mathcal{D}_k^+}[\phi(x; \theta_{init})]$
    
    \Statex \hrulefill \vspace{2pt} \textit{ // Defined in Fig. 1 (Top Right)}
    \Function{Mix}{$z_i, z_j, \lambda$}
        \If{$\mathcal{G} == \textsc{Slerp}$} \Comment{\textcolor{geoPrimary}{\textbf{Geodesic Path}}}
            \State $\Omega \leftarrow \arccos(z_i \cdot z_j)$
            \State \textbf{return} $\frac{\sin((1-\lambda)\Omega)}{\sin(\Omega)}z_i + \frac{\sin(\lambda\Omega)}{\sin(\Omega)}z_j$
        \Else \Comment{\textcolor{linearColor}{\textbf{Linear Chord}}}
            \State \textbf{return} $\lambda z_i + (1-\lambda)z_j$
        \EndIf
    \EndFunction

    \Statex \hrulefill \vspace{2pt} \textit{ // Local Training Objective (Eq. \ref{eq:local_opt})}
    \Function{ClientUpdate}{$\theta, \mathcal{D}_k^+$}
        \For{$e = 1 \dots E$}
            \For{batch $x \sim \mathcal{D}_k^+$}
                \State $z \leftarrow \phi(x; \theta) = \hat{z} / \|\hat{z}\|_2$, \quad $\hat{z} = \text{Enc}(x; \theta)$
                \State $z_{pair} \leftarrow \text{Permute}(z)$, sample $\lambda \sim \text{Beta}(\kappa, \kappa)$
                \State $\tilde{z} \leftarrow \Call{Mix}{z, z_{pair}, \lambda}$
                \State $\mathcal{L} \leftarrow \| \tilde{z} - \mathbf{c} \|^2$ \Comment{\textbf{Single-Center Loss}}
                \State $\theta \leftarrow \theta - \eta \nabla_\theta \mathcal{L}$
            \EndFor
        \EndFor
        \State \textbf{return} $\theta$
    \EndFunction

    \Statex \hrulefill \vspace{2pt} \textit{ // Unified Global Inference (Eq. \ref{eq:global_obj})}
    \Function{NLL}{$x, \theta, \mathbf{c}$}
        \State $d^2 \leftarrow \|\phi(x; \theta) - \mathbf{c}\|^2$
        \State \textbf{return} $\log(1 + d^2)$ \textbf{\Comment{Neg. Log-Likelihood (Student-t)}}
    \EndFunction

    \If{$\mathcal{P} == \textsc{OneShot}$} 
        \State \textbf{Parallel:} $\theta_k \leftarrow \Call{ClientUpdate}{\theta_{init}, \mathcal{D}_k^+}$
        \State \textbf{Ensemble:} $\Theta \leftarrow \{\theta_1 \dots \theta_K\}$
        \State \textbf{Score:} $\mathcal{A}(x) \leftarrow \sum_{\theta \in \Theta} \Call{NLL}{x, \theta, \mathbf{c}}$ \textbf{\Comment{Sum of NLLs}}
    \Else 
        \For{$t = 1 \dots T$}
            \State \textbf{Parallel:} $\theta_k^t \leftarrow \Call{ClientUpdate}{\theta_g, \mathcal{D}_k^+}$
            \State \textbf{Server:} $\theta_g \leftarrow \sum \frac{N_k}{N} \theta_k^t$ 
        \EndFor
        \State \textbf{Score:} $\mathcal{A}(x) \leftarrow \Call{NLL}{x, \theta_g, \mathbf{c}}$ \textbf{\Comment{Single Model NLL}}
    \EndIf
    \State \textbf{Decision:} \textbf{return} $1$ if $\mathcal{A}(x) > \tau$ else $0$

\end{algorithmic}
\end{algorithm}

\section{Experiments}
\label{sec:experiments}
We conducted a comprehensive empirical evaluation to test \textit{Geodesic-SVDD} in different distributed tabular scenarios. 

\subsection{Experimental Setup}
\subsubsection{\textbf{Datasets}}
We utilized five standard tabular benchmarks representing diverse complexities: \textit{Iris}~\cite{iris53}, \textit{Breast Cancer}~\cite{174969}, \textit{Titanic}~\cite{kaggle_titanic}, \textit{Adult}~\cite{adult_2}, and \textit{Covertype}~\cite{covertype_31}. \textit{Covertype} is the largest and most complex benchmark (581{,}012 samples, 54 features, 7 classes) and serves as a stress test for manifold learning. Features were processed via \textit{StandardScaler}, except for Titanic and Adult which utilized \textit{RobustScaler} to handle outliers in skewed numerical features.

\subsubsection{\textbf{Non-IID Setting}}
We simulate a Cross-Silo setting with $K=3$ organizations. To model non-IID distributions, we partition data using a Dirichlet distribution $\text{Dir}(\alpha)$ over the label space. We evaluate four increasingly difficult scenarios: \textit{IID} (Uniform), \textit{Mild Skew} ($\alpha=0.5$), \textit{Moderate Skew} ($\alpha=0.1$), and \textit{Extreme Skew} ($\alpha=0.01$). In the $\alpha=0.01$ case, clients possess near-mutually exclusive classes, representing the ``disjoint support'' defined in Eq. (1).

\subsubsection{\textbf{Baselines \& Metrics}}
We compare against:
\begin{itemize}
    \item \textbf{Centralized SVDD~\cite{ruff2018deep} / k-NN~\cite{cover1967nearest}:} Reference models trained on pooled (IID) data, using Deep SVDD for manifold density and k-Nearest Neighbors for non-parametric density estimation.
    \item \textbf{FedAvg:} To separate the performance benefits of our architecture from our training strategy, we evaluate two variants of this baseline. The first employs a standard MLP as the feature extractor, while the second utilizes the \textit{exact feature extractor} of our proposed method. Both optimize the spherical SVDD objective~\cite{ruff2018deep} via standard weight averaging~\cite{mcmahan2017communication}, allowing to isolate gains derived specifically from our aggregation method versus the SVDD loss.
    \item \textbf{Fed-Forest:} A federated Isolation Forest~\cite{liu2020federated} in which each client trains isolation trees locally and the server pools them into a global ensemble.
        \item \textbf{Fed-XGB:} each client trains XGBoost~\cite{chen2016xgboost} to separate its local data from locally generated synthetic negatives (feature-wise permutations and uniform samples); the server averages the clients' scores (our implementation)
\end{itemize}
We adopt a \textbf{One-vs-All} evaluation protocol. For a dataset with classes $\mathcal{C}$, we conduct $|\mathcal{C}|$ distinct experiments. In each experiment, a single class $c \in \mathcal{C}$ is designated as the \textit{normal} class, and the models are trained exclusively on samples where $y=c$ (one-class setting). During evaluation, samples from class $c$ are labeled negative (0), while samples from all other classes ($y \neq c$) are labeled positive (anomalous, 1). We report the macro-average of \textbf{AUROC}, Accuracy, Precision, Recall, and F1 across these experiments. Threshold-dependent metrics are computed at the optimal operating point determined via Youden's J statistic~\cite{ruopp2008youden} on a held-out server-side validation set containing both in-distribution and out-of-distribution samples. Results are averaged over 5 independent seeded runs ($\text{mean} \pm \text{std}$). Unlike supervised baselines that leverage contrasting labels to learn discriminative boundaries, our method operates as a \textbf{One-Class Classifier}. We solve the strictly harder problem of \textit{density estimation} without access to negative samples during local training.

\subsection{Results}
\label{sec:comparison_sota}
\begin{figure}[t]
    \centering
    \includegraphics[width=0.85\columnwidth]{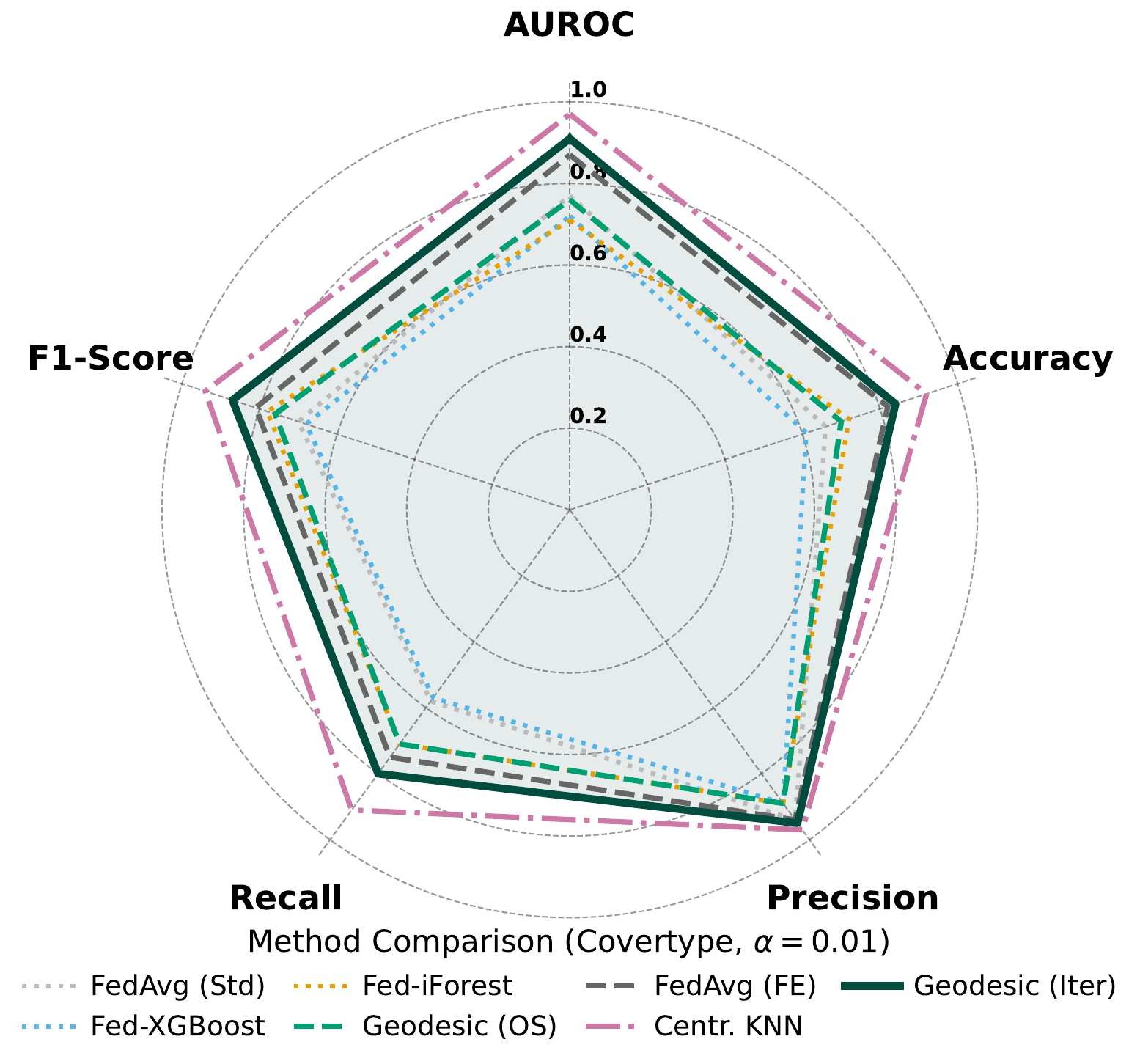}
    \caption{\textbf{Performance Profile (Covertype, $\alpha=0.01$).} A radar chart comparison of Geodesic-SVDD against the federated baselines; centralized k-NN (IID) is shown for reference. Our method maintains balanced precision and recall, whereas Fed-XGB and FedAvg (SM) fall below $0.60$ recall.}
    \label{fig:radar_chart}
\end{figure}

We compare Geodesic-SVDD against federated tree ensembles (Fed-Forest, Fed-XGB) and three FedAvg variants (Tables~\ref{tab:results_part1} and~\ref{tab:results_part2}). Tree ensembles are competitive only on low-dimensional data: Fed-Forest reaches $0.96$--$0.97$ AUROC on \textit{Iris} and leads on \textit{Adult}, where all methods except Fed-XGB lie between $0.58$ and $0.66$ AUROC with standard deviations of up to 20 points, but it plateaus at $0.71$ on \textit{Covertype}. Fed-XGB, which relies on locally generated synthetic negatives, is the weakest method overall, with $0.72$ AUROC on \textit{Covertype} at $\alpha=0.01$ and below $0.50$ on \textit{Adult}.
\textit{Geodesic-SVDD (Iter)} is the most consistent method across datasets and skew levels: it attains the best or second-best score in 57 of the 100 metric columns, more than any baseline (FedAvg (No SM): 40). On \textit{Covertype} at $\alpha=0.01$, it reaches $0.91$ AUROC, 18--20 points above both tree ensembles, with a recall of $0.80$ versus $0.71$ (Fed-Forest) and $0.57$ (Fed-XGB) (Fig.~\ref{fig:radar_chart}). On \textit{Titanic}, it is the best federated method at every skew level ($0.75$ AUROC). The strongest baseline, FedAvg (No SM), leads on \textit{Iris} and matches Geodesic-SVDD (Iter) on \textit{Covertype} to within one AUROC point. Notably, FedAvg (FE), which shares our encoder but not our geodesic mixup, drops from $0.91$ to $0.85$ AUROC on \textit{Covertype} at $\alpha=0.1$, whereas Geodesic-SVDD (Iter) holds $0.91$, suggesting that geodesic latent mixup stabilises training under label skew.

\begin{figure*}[t!]
    \centering
    \includegraphics[width=\textwidth]{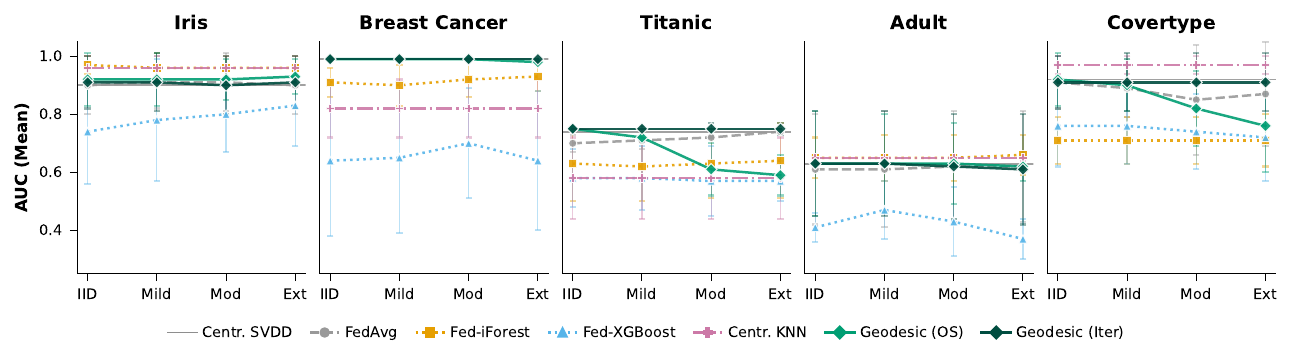}
    \caption{\textbf{Resilience to Non-IID Label Skew.} Performance (AUC) across five benchmarks with increasing non-IID partition skew. Centralized references are trained on pooled IID data and drawn as flat lines.}
    \label{fig:nature_trend}
\end{figure*}

We analyzed the operational trade-off between the privacy-preserving \textit{One-Shot (OSFL)} protocol and standard \textit{Iterative Averaging}.
As shown in Table~\ref{tab:results_part1} and \ref{tab:results_part2}, the OSFL protocol ($R=1$) matches centralized Deep SVDD under IID and mild skew ($\alpha=0.5$), with a mean AUROC of $0.84$ and $0.83$ versus $0.84$. On \textit{Breast Cancer}, for example, OSFL attains $0.99$ AUROC, on par with centralized training. This makes OSFL a strong choice for high-security environments where bi-directional communication increases the attack surface.
 Figure \ref{fig:nature_trend} illustrates the performance trajectory as heterogeneity increases. While tree baselines remain flat but low (e.g., $0.71$ for Fed-Forest on \textit{Covertype}), the One-Shot approach degrades under strong skew, dropping from $0.92$ (IID) to $0.76$ ($\alpha=0.01$) on \textit{Covertype} and from $0.75$ to $0.59$ on \textit{Titanic}. Switching to the \textbf{Iterative} protocol recovers this loss, maintaining $0.91$ and $0.75$ AUROC, respectively, even under extreme skew. This establishes a clear hierarchy: employ OSFL for bandwidth efficiency in low-skew settings, and Iterative Averaging for partition alignment in high-skew regimes.

\subsection{Statistical Significance Analysis}
\label{sec:stats}

\begin{figure}[h]
    \centering
    \includegraphics[width=0.9\columnwidth]{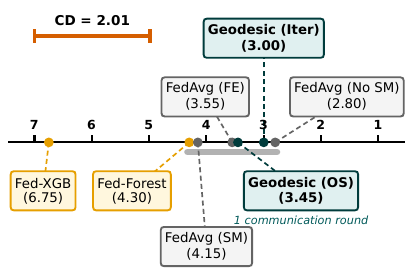}
\caption{\textbf{Critical Difference Diagram (Friedman--Nemenyi, $\alpha=0.05$).} Average AUROC ranks over 20 scenarios (lower is better). Both Geodesic-SVDD variants rank in the top three; only Fed-XGB is significantly worse ($CD=2.01$).}
\label{fig:cd_diagram}
\end{figure}

To assess robustness beyond individual datasets, we rank the seven federated methods by seed-averaged AUROC over $N=20$ scenarios (5 datasets $\times$ 4 skew levels); centralized references are excluded, as they have no non-IID counterpart. The \textit{Friedman test} rejects equal performance ($\chi^2_F=45.6$, $p<10^{-7}$), and the \textit{Nemenyi post-hoc test} ($\alpha=0.05$) yields $CD=2.01$~\cite{JMLRdemsar06a} (Fig.~\ref{fig:cd_diagram}).

\textit{Geodesic-SVDD (Iter)} achieves the highest mean AUROC across all scenarios (83.5) and ranks second (3.00), statistically tied with FedAvg (No SM) (2.80) and ahead of every other baseline. Since rank tests count sub-point gaps as full wins, we also apply the Bayesian signed-rank test~\cite{benavoli2017time} with a region of practical equivalence of $\pm 1$ AUROC point: Geodesic-SVDD (Iter) is practically superior to FedAvg (SM), Fed-Forest and Fed-XGB ($P \geq 0.98$), never inferior to FedAvg (FE) ($P < 0.01$), and has the lowest worst-case probability of being outperformed by any competitor (0.18, vs.\ 0.34 for FedAvg (No SM)). Crucially, \textit{Geodesic-SVDD (OS)} ranks third (3.45) with a \emph{single} communication round, within the critical difference of the best iterative methods; its gap to the iterative protocol is not significant and concentrates at high skew ($\alpha \leq 0.1$, Table~\ref{tab:results_part2}).

 \subsection{Efficiency vs. Accuracy Trade-off}
\label{sec:efficiency}
While predictive performance is paramount, one of the primary motivations for the One-Shot protocol is communication efficiency in bandwidth-constrained settings. We quantify the communication cost relative to the model size $M$ (in MB).
Standard \texttt{FedAvg} and Iterative Geodesic-SVDD require $2 \times R$ transmissions (upload/download) per client, resulting in a total communication cost of $\mathcal{C}_{iter} = 2 \cdot R \cdot K \cdot M$. In contrast, our One-Shot protocol requires a single upload, reducing cost to $\mathcal{C}_{OS} = K \cdot M$. We exclude the warm-up phase communication overhead from this analysis, as transmitting a single $d$-dimensional latent statistics vector ($\mathbb{E}[\phi(x)]$) is asymptotically negligible compared to the full model size $M$.
For a standard training session of $R=50$ rounds, Geodesic-SVDD (OS) reduces communication overhead by a factor of $100\times$. 
As seen in Table~\ref{tab:results_part1}, on the \textit{Breast Cancer} dataset, this $100\times$ reduction in bandwidth costs no AUROC under IID ($0.992$ vs.\ $0.991$ for the iterative protocol) and at most $1.2$ points under extreme skew ($0.978$ vs.\ $0.990$).
However, on complex manifolds like \textit{Covertype} (Fig.~\ref{fig:nature_trend}), iterative alignment is necessary at high skew: at $\alpha=0.01$, it recovers 15 of the 16 AUROC points lost by the One-Shot protocol ($0.76 \to 0.91$).

\subsection{Ablation: The Geometry of Latent Interpolation}
\label{sec:ablation}

Finally, we isolate the impact of our geometric constraints by comparing Spherical (\texttt{SLERP}) against standard Linear (\texttt{LERP}) interpolation. Table~\ref{tab:ablation_study} presents a compelling geometric paradox: on downstream predictive tasks, \textit{Linear-SVDD} achieves scores (AUROC $0.920$) statistically indistinguishable from \textit{Geodesic-SVDD} ($0.921$). From a purely utilitarian perspective, the linear approximation appears sufficient. However, a deeper structural analysis reveals that Linear Mixup achieves this via ``lucky'' regularization, masking two critical failures on the $\mathbb{S}^{d-1}$ manifold that we quantify as follows:

\begin{itemize}
    \item \textbf{Protocol 1: Manifold Collapse ($d=128$).} 
    To quantify topological violation, we sampled $N=1000$ pairs from an isotropic distribution on the hypersphere ($z \sim \mathcal{U}(\mathbb{S}^{d-1})$) derived from standard normal vectors $z_{\text{raw}} \sim \mathcal{N}(0, I_d)$. We measured the Euclidean norm of the interpolated samples at the midpoint ($\lambda=0.5$). Linear paths collapsed the latent norm to $\|\tilde{z}\| \approx 0.705 \pm 0.03$, systematically projecting valid data into the low-density interior (the SVDD ``anomaly zone''), whereas SLERP preserves $\|\tilde{z}\| \equiv 1.0$. 
    
    \item \textbf{Protocol 2: Angular Velocity Instability ($10^9\times$).} 
    To quantify optimization stability, we measured the angular velocity variance ($\sigma^2_{\omega} = \text{Var}[\frac{\partial \Omega}{\partial \lambda}]$) along the interpolation path $\lambda \in [0,1]$. We simulated trajectories between semi-antipodal feature vectors ($\theta \approx 160^\circ$). While Geodesic interpolation maintains constant velocity ($\sigma^2_{\text{geo}} \approx 5.5 \times 10^{-12}$), Linear interpolation stagnates near the poles and accelerates violently at the midpoint. This results in a velocity variance of $\sigma^2_{\text{lin}} \approx 3.99 \times 10^{-3}$, representing a $10^9\times$ instability gap in the geometric gradient.
\end{itemize}

Consequently, while Linear Mixup works on static benchmarks where data is locally clustered, \texttt{SLERP} is essential to ground high detection performance in mathematically stable boundary definitions rather than distributional artifacts.

\begin{table*}[!t]
\centering
\caption{Comparative Performance (Part I): IID and Mild Non-IID Scenarios. \\
\textit{Mean $\pm$ Std. Dev. (in \%). Best federated result per column in \textbf{bold}, second best \underline{underlined}; centralized baselines are not ranked. \colorbox{geoPrimary!10}{Teal rows} indicate our method.}}
\label{tab:results_part1}
\resizebox{\textwidth}{!}{%
\begin{tabular}{ll ccccc | ccccc}
\toprule
\multirow{2}{*}{\textbf{Dataset}} & \multirow{2}{*}{\textbf{Method}} 
& \multicolumn{5}{c}{\textbf{IID (Uniform)}} 
& \multicolumn{5}{c}{\textbf{Dirichlet ($\alpha=0.5$) [Mild]}} \\
\cmidrule(lr){3-7} \cmidrule(lr){8-12} 
& & \textbf{AUC} & \textbf{Acc} & \textbf{Prec} & \textbf{Rec} & \textbf{F1} & \textbf{AUC} & \textbf{Acc} & \textbf{Prec} & \textbf{Rec} & \textbf{F1} \\
\midrule

\multirow{9}{*}{\rotatebox{90}{\textbf{Iris}}}
& \cellcolor{gray!10}Centralized SVDD & \cellcolor{gray!10}90.10$\pm$11.0 & \cellcolor{gray!10}82.70$\pm$14.8 & \cellcolor{gray!10}91.80$\pm$13.7 & \cellcolor{gray!10}83.20$\pm$15.3 & \cellcolor{gray!10}86.00$\pm$12.3 & \multicolumn{5}{c}{\cellcolor{gray!10}\textit{-- Centralized Baselines --}} \\
& \cellcolor{gray!10}Centralized k-NN & \cellcolor{gray!10}96.00$\pm$4.00 & \cellcolor{gray!10}88.00$\pm$8.00 & \cellcolor{gray!10}97.00$\pm$4.00 & \cellcolor{gray!10}87.00$\pm$6.00 & \cellcolor{gray!10}91.00$\pm$6.00 & \multicolumn{5}{c}{\cellcolor{gray!10}\textit{}} \\
& FedAvg (FE) & 90.40$\pm$10.1 & 82.00$\pm$13.6 & 91.10$\pm$8.40 & 82.70$\pm$12.1 & 85.50$\pm$11.6 & 91.30$\pm$8.80 & 82.40$\pm$13.6 & 93.40$\pm$5.20 & 81.10$\pm$15.4 & 84.90$\pm$12.9 \\
& FedAvg (No SM) & \textbf{98.32$\pm$1.85} & \textbf{93.33$\pm$4.84} & \textbf{96.71$\pm$2.37} & \textbf{93.15$\pm$6.16} & \textbf{94.58$\pm$4.21} & \textbf{98.20$\pm$1.89} & \underline{92.44$\pm$5.29} & \underline{96.74$\pm$2.34} & \underline{91.86$\pm$6.74} & \underline{93.73$\pm$4.83} \\
& FedAvg (SM) & 96.84$\pm$2.50 & 88.89$\pm$6.45 & 95.31$\pm$3.54 & 87.77$\pm$6.34 & 90.91$\pm$5.27 & 96.08$\pm$2.85 & 87.56$\pm$7.39 & \textbf{96.79$\pm$2.41} & 84.42$\pm$8.81 & 89.56$\pm$6.22 \\
& Fed-Forest & \underline{96.99$\pm$3.09} & \underline{90.89$\pm$6.72} & 95.24$\pm$5.67 & \underline{91.34$\pm$9.70} & \underline{92.83$\pm$5.70} & \underline{96.18$\pm$4.69} & \textbf{92.89$\pm$4.37} & 95.43$\pm$4.01 & \textbf{94.07$\pm$5.67} & \textbf{94.61$\pm$3.35} \\
& Fed-XGB & 73.77$\pm$18.1 & 74.89$\pm$12.6 & 83.24$\pm$12.3 & 77.07$\pm$22.1 & 78.28$\pm$16.7 & 77.94$\pm$21.2 & 74.00$\pm$15.9 & 81.05$\pm$24.3 & 75.96$\pm$26.8 & 76.30$\pm$23.0 \\
\cmidrule{2-12}
\rowcolor{geoPrimary!10} & \textbf{Geodesic (OS)} & 91.60$\pm$8.50 & 83.30$\pm$11.6 & \underline{95.40$\pm$4.30} & 80.10$\pm$13.1 & 85.70$\pm$10.4 & 91.50$\pm$8.50 & 84.20$\pm$10.6 & 96.10$\pm$2.90 & 80.20$\pm$13.9 & 86.00$\pm$10.2 \\
\rowcolor{geoPrimary!10} & \textbf{Geodesic (Iter)} & 90.50$\pm$9.20 & 83.10$\pm$12.5 & 94.80$\pm$4.80 & 81.10$\pm$13.3 & 85.30$\pm$11.8 & 90.60$\pm$9.70 & 84.40$\pm$13.0 & 94.80$\pm$4.90 & 82.60$\pm$14.2 & 86.30$\pm$12.5 \\
\midrule

\multirow{9}{*}{\rotatebox{90}{\textbf{BreastC}}}
& \cellcolor{gray!10}Centralized SVDD & \cellcolor{gray!10}99.00$\pm$0.70 & \cellcolor{gray!10}95.70$\pm$2.50 & \cellcolor{gray!10}96.30$\pm$1.20 & \cellcolor{gray!10}94.60$\pm$5.20 & \cellcolor{gray!10}95.40$\pm$2.60 & \multicolumn{5}{c}{\cellcolor{gray!10}\textit{-- Centralized Baselines --}} \\
& \cellcolor{gray!10}Centralized k-NN & \cellcolor{gray!10}82.00$\pm$10.0 & \cellcolor{gray!10}77.00$\pm$6.00 & \cellcolor{gray!10}75.00$\pm$4.00 & \cellcolor{gray!10}83.00$\pm$4.00 & \cellcolor{gray!10}78.00$\pm$1.00 & \multicolumn{5}{c}{\cellcolor{gray!10}\textit{}} \\
& FedAvg (FE) & \textbf{99.20$\pm$0.00} & \textbf{96.40$\pm$0.80} & \textbf{97.20$\pm$0.80} & \textbf{95.20$\pm$1.80} & \textbf{96.10$\pm$0.60} & \textbf{99.10$\pm$0.00} & \underline{96.40$\pm$0.80} & \textbf{97.00$\pm$0.30} & \textbf{95.50$\pm$1.40} & \textbf{96.20$\pm$0.50} \\
& FedAvg (No SM) & 98.65$\pm$0.20 & 94.56$\pm$1.05 & 94.32$\pm$1.44 & 94.24$\pm$0.31 & 94.16$\pm$0.93 & 98.41$\pm$0.03 & 93.86$\pm$0.70 & 93.06$\pm$2.28 & 93.63$\pm$0.76 & 93.25$\pm$1.54 \\
& FedAvg (SM) & 98.71$\pm$0.01 & 95.00$\pm$0.79 & 95.70$\pm$0.69 & 93.86$\pm$0.99 & 94.65$\pm$0.88 & \underline{98.83$\pm$0.07} & 95.26$\pm$1.40 & 96.58$\pm$0.88 & 93.79$\pm$0.70 & 95.09$\pm$0.11 \\
& Fed-Forest & 91.41$\pm$5.48 & 85.18$\pm$5.09 & 85.68$\pm$7.48 & 83.74$\pm$6.47 & 84.30$\pm$4.61 & 89.89$\pm$7.06 & 83.68$\pm$5.74 & 83.59$\pm$8.46 & 83.41$\pm$5.43 & 83.14$\pm$4.87 \\
& Fed-XGB & 64.36$\pm$26.2 & 68.16$\pm$19.6 & 69.17$\pm$7.68 & 75.42$\pm$35.5 & 65.03$\pm$27.5 & 64.51$\pm$26.4 & 65.61$\pm$18.5 & 57.62$\pm$24.8 & 64.57$\pm$34.9 & 58.45$\pm$29.4 \\
\cmidrule{2-12}
\rowcolor{geoPrimary!10} & \textbf{Geodesic (OS)} & \textbf{99.20$\pm$0.10} & \underline{96.10$\pm$0.10} & 96.60$\pm$0.30 & 94.40$\pm$3.70 & 95.30$\pm$1.80 & 98.60$\pm$0.30 & 95.30$\pm$0.90 & 95.00$\pm$1.20 & 93.00$\pm$4.20 & 93.90$\pm$2.80 \\
\rowcolor{geoPrimary!10} & \textbf{Geodesic (Iter)} & \underline{99.10$\pm$0.00} & \underline{96.10$\pm$0.50} & \underline{96.80$\pm$0.20} & \underline{94.80$\pm$2.00} & \underline{95.70$\pm$1.00} & \textbf{99.10$\pm$0.00} & \textbf{96.50$\pm$0.40} & \underline{96.90$\pm$0.00} & \underline{95.30$\pm$1.90} & \underline{96.00$\pm$1.00} \\
\midrule

\multirow{9}{*}{\rotatebox{90}{\textbf{Titanic}}}
& \cellcolor{gray!10}Centralized SVDD & \cellcolor{gray!10}74.40$\pm$2.60 & \cellcolor{gray!10}71.80$\pm$2.10 & \cellcolor{gray!10}68.80$\pm$8.50 & \cellcolor{gray!10}75.40$\pm$8.90 & \cellcolor{gray!10}71.70$\pm$7.40 & \multicolumn{5}{c}{\cellcolor{gray!10}\textit{-- Centralized Baselines --}} \\
& \cellcolor{gray!10}Centralized k-NN & \cellcolor{gray!10}58.00$\pm$14.0 & \cellcolor{gray!10}58.00$\pm$12.0 & \cellcolor{gray!10}65.00$\pm$2.00 & \cellcolor{gray!10}45.00$\pm$14.0 & \cellcolor{gray!10}46.00$\pm$14.0 & \multicolumn{5}{c}{\cellcolor{gray!10}\textit{}} \\
& FedAvg (FE) & \underline{69.90$\pm$3.10} & 65.00$\pm$3.70 & 67.80$\pm$7.40 & 61.00$\pm$6.90 & 62.50$\pm$0.90 & 70.50$\pm$3.00 & 66.70$\pm$3.20 & \textbf{69.90$\pm$8.60} & 61.80$\pm$7.40 & 63.70$\pm$0.50 \\
& FedAvg (No SM) & 64.05$\pm$10.7 & 61.87$\pm$9.20 & 69.53$\pm$4.56 & 46.42$\pm$13.0 & 53.58$\pm$8.35 & 64.52$\pm$10.4 & 61.83$\pm$8.02 & 67.04$\pm$5.09 & 50.95$\pm$12.9 & 55.95$\pm$6.79 \\
& FedAvg (SM) & 65.44$\pm$0.87 & 62.06$\pm$1.91 & 62.01$\pm$7.21 & 61.81$\pm$0.92 & 58.98$\pm$3.41 & 63.13$\pm$2.18 & 60.53$\pm$1.98 & 59.91$\pm$6.64 & 64.99$\pm$1.97 & 58.75$\pm$4.53 \\
& Fed-Forest & 63.11$\pm$12.5 & 57.94$\pm$11.8 & 63.01$\pm$8.89 & 46.47$\pm$27.3 & 47.63$\pm$20.3 & 62.41$\pm$12.2 & 56.34$\pm$11.3 & 64.08$\pm$7.41 & 38.47$\pm$23.3 & 42.49$\pm$18.5 \\
& Fed-XGB & 57.86$\pm$9.89 & 56.22$\pm$9.64 & 64.39$\pm$9.48 & 40.15$\pm$20.7 & 44.40$\pm$15.1 & 57.91$\pm$11.1 & 55.80$\pm$10.3 & 66.62$\pm$13.0 & 36.63$\pm$20.9 & 41.17$\pm$18.1 \\
\cmidrule{2-12}
\rowcolor{geoPrimary!10} & \textbf{Geodesic (OS)} & \textbf{74.90$\pm$1.40} & \textbf{71.50$\pm$0.50} & \textbf{70.10$\pm$7.90} & \underline{72.20$\pm$2.70} & \underline{70.70$\pm$5.20} & \underline{71.90$\pm$1.20} & \underline{68.90$\pm$1.20} & 68.60$\pm$10.8 & \underline{72.10$\pm$3.10} & \underline{69.10$\pm$4.20} \\
\rowcolor{geoPrimary!10} & \textbf{Geodesic (Iter)} & \textbf{74.90$\pm$1.40} & \underline{71.20$\pm$0.80} & \underline{69.80$\pm$8.90} & \textbf{73.10$\pm$0.50} & \textbf{70.90$\pm$4.90} & \textbf{75.10$\pm$1.20} & \textbf{71.60$\pm$1.60} & \underline{69.80$\pm$9.00} & \textbf{73.40$\pm$2.90} & \textbf{71.10$\pm$6.20} \\
\midrule

\multirow{9}{*}{\rotatebox{90}{\textbf{Adult}}}
& \cellcolor{gray!10}Centralized SVDD & \cellcolor{gray!10}62.90$\pm$17.6 & \cellcolor{gray!10}68.90$\pm$1.40 & \cellcolor{gray!10}61.40$\pm$30.6 & \cellcolor{gray!10}44.50$\pm$21.1 & \cellcolor{gray!10}51.50$\pm$25.0 & \multicolumn{5}{c}{\cellcolor{gray!10}\textit{-- Centralized Baselines --}} \\
& \cellcolor{gray!10}Centralized k-NN & \cellcolor{gray!10}65.00$\pm$1.00 & \cellcolor{gray!10}74.00$\pm$5.00 & \cellcolor{gray!10}75.00$\pm$7.00 & \cellcolor{gray!10}52.00$\pm$22.0 & \cellcolor{gray!10}60.00$\pm$18.0 & \multicolumn{5}{c}{\cellcolor{gray!10}\textit{}} \\
& FedAvg (FE) & 61.00$\pm$18.9 & 68.70$\pm$0.10 & 60.10$\pm$32.5 & 40.00$\pm$23.6 & 47.70$\pm$27.7 & 60.80$\pm$19.6 & \underline{69.50$\pm$0.00} & 59.80$\pm$32.7 & 39.70$\pm$25.3 & 47.40$\pm$28.9 \\
& FedAvg (No SM) & 62.53$\pm$18.0 & \underline{70.08$\pm$0.67} & 60.77$\pm$32.0 & 39.10$\pm$25.5 & 47.18$\pm$28.9 & 62.55$\pm$18.1 & 69.30$\pm$0.12 & 60.25$\pm$32.5 & 40.05$\pm$24.2 & 47.98$\pm$27.9 \\
& FedAvg (SM) & 60.08$\pm$0.30 & 54.36$\pm$4.69 & 61.54$\pm$26.8 & \textbf{51.83$\pm$12.9} & 47.49$\pm$4.49 & \underline{64.80$\pm$1.99} & 61.58$\pm$2.19 & \underline{63.00$\pm$25.8} & \textbf{55.37$\pm$2.13} & \underline{53.43$\pm$12.6} \\
& Fed-Forest & \textbf{65.31$\pm$7.48} & \textbf{70.64$\pm$2.11} & \underline{63.91$\pm$22.5} & \underline{51.80$\pm$18.5} & \textbf{57.15$\pm$20.3} & \textbf{64.83$\pm$7.64} & \textbf{70.26$\pm$1.47} & 61.94$\pm$23.8 & \underline{53.56$\pm$19.2} & \textbf{57.37$\pm$21.3} \\
& Fed-XGB & 40.86$\pm$4.51 & 52.67$\pm$23.4 & \textbf{66.51$\pm$18.2} & 10.44$\pm$5.40 & 17.34$\pm$8.43 & 46.88$\pm$10.1 & 54.59$\pm$19.1 & \textbf{63.48$\pm$22.0} & 28.42$\pm$28.4 & 30.50$\pm$19.5 \\
\cmidrule{2-12}
\rowcolor{geoPrimary!10} & \textbf{Geodesic (OS)} & \underline{63.30$\pm$16.6} & 68.90$\pm$1.40 & 61.60$\pm$30.0 & 46.20$\pm$19.7 & \underline{52.60$\pm$24.0} & 63.30$\pm$16.7 & 68.90$\pm$0.60 & 61.70$\pm$29.9 & 46.00$\pm$19.9 & 52.50$\pm$24.1 \\
\rowcolor{geoPrimary!10} & \textbf{Geodesic (Iter)} & 62.60$\pm$17.7 & 69.20$\pm$0.30 & 61.70$\pm$30.8 & 42.20$\pm$21.9 & 50.00$\pm$25.7 & 62.60$\pm$17.6 & 69.20$\pm$0.50 & 61.60$\pm$31.1 & 42.10$\pm$21.4 & 50.00$\pm$25.3 \\
\midrule

\multirow{9}{*}{\rotatebox{90}{\textbf{Covertype}}}
& \cellcolor{gray!10}Centralized SVDD & \cellcolor{gray!10}92.10$\pm$9.20 & \cellcolor{gray!10}85.60$\pm$10.7 & \cellcolor{gray!10}94.40$\pm$9.20 & \cellcolor{gray!10}83.10$\pm$14.4 & \cellcolor{gray!10}88.20$\pm$12.1 & \multicolumn{5}{c}{\cellcolor{gray!10}\textit{-- Centralized Baselines --}} \\
& \cellcolor{gray!10}Centralized k-NN & \cellcolor{gray!10}97.00$\pm$3.00 & \cellcolor{gray!10}92.00$\pm$5.00 & \cellcolor{gray!10}97.00$\pm$5.00 & \cellcolor{gray!10}91.00$\pm$5.00 & \cellcolor{gray!10}94.00$\pm$5.00 & \multicolumn{5}{c}{\cellcolor{gray!10}\textit{}} \\
& FedAvg (FE) & 91.00$\pm$9.70 & 84.50$\pm$11.1 & 94.60$\pm$9.00 & 81.20$\pm$15.8 & 87.00$\pm$13.1 & 88.80$\pm$13.3 & 83.40$\pm$13.2 & \textbf{95.70$\pm$6.70} & 77.50$\pm$23.1 & 83.70$\pm$19.8 \\
& FedAvg (No SM) & \underline{91.45$\pm$8.78} & 84.42$\pm$10.3 & \underline{94.72$\pm$8.53} & 81.43$\pm$14.5 & 87.33$\pm$12.0 & \textbf{91.70$\pm$9.25} & \textbf{85.14$\pm$10.8} & \underline{94.86$\pm$8.35} & \textbf{81.92$\pm$15.4} & \textbf{87.61$\pm$12.6} \\
& FedAvg (SM) & 75.90$\pm$13.5 & 65.83$\pm$15.1 & \textbf{94.78$\pm$8.74} & 58.09$\pm$22.5 & 69.31$\pm$20.1 & 78.21$\pm$13.2 & 68.84$\pm$13.7 & 93.15$\pm$10.9 & 62.77$\pm$20.0 & 73.08$\pm$18.1 \\
& Fed-Forest & 71.14$\pm$8.23 & 69.77$\pm$10.2 & 88.86$\pm$15.4 & 68.87$\pm$12.2 & 77.41$\pm$13.1 & 70.51$\pm$7.81 & 71.15$\pm$11.6 & 88.84$\pm$15.3 & 70.68$\pm$13.3 & 78.42$\pm$13.5 \\
& Fed-XGB & 76.30$\pm$13.9 & 65.66$\pm$12.1 & 90.27$\pm$15.1 & 58.89$\pm$20.7 & 70.34$\pm$20.1 & 75.76$\pm$12.5 & 65.68$\pm$11.0 & 91.13$\pm$13.6 & 59.38$\pm$18.7 & 71.12$\pm$18.0 \\
\cmidrule{2-12}
\rowcolor{geoPrimary!10} & \textbf{Geodesic (OS)} & \textbf{92.00$\pm$8.80} & \textbf{85.10$\pm$10.4} & 94.30$\pm$9.50 & \textbf{82.90$\pm$13.5} & \textbf{88.00$\pm$11.7} & 90.40$\pm$8.70 & 83.50$\pm$10.2 & 94.10$\pm$9.60 & 81.20$\pm$13.4 & 87.00$\pm$11.6 \\
\rowcolor{geoPrimary!10} & \textbf{Geodesic (Iter)} & 91.10$\pm$9.30 & \underline{84.60$\pm$10.7} & 94.20$\pm$9.60 & \underline{82.10$\pm$14.4} & \underline{87.50$\pm$12.4} & \underline{91.30$\pm$9.50} & \underline{84.70$\pm$10.9} & 94.50$\pm$9.20 & \underline{81.70$\pm$15.2} & \underline{87.30$\pm$12.7} \\
\bottomrule
\end{tabular}
}
\end{table*}

\begin{table*}[!t]
\centering
\caption{Comparative Performance (Part II): Moderate and Extreme Non-IID Scenarios. \\
\textit{Same format as Table~\ref{tab:results_part1}: best federated result per column in \textbf{bold}, second best \underline{underlined}.}}
\label{tab:results_part2}
\resizebox{\textwidth}{!}{%
\begin{tabular}{ll ccccc | ccccc}
\toprule
\multirow{2}{*}{\textbf{Dataset}} & \multirow{2}{*}{\textbf{Method}} 
& \multicolumn{5}{c}{\textbf{Dirichlet ($\alpha=0.1$) [Moderate]}} 
& \multicolumn{5}{c}{\textbf{Dirichlet ($\alpha=0.01$) [Extreme]}} \\
\cmidrule(lr){3-7} \cmidrule(lr){8-12} 
& & \textbf{AUC} & \textbf{Acc} & \textbf{Prec} & \textbf{Rec} & \textbf{F1} & \textbf{AUC} & \textbf{Acc} & \textbf{Prec} & \textbf{Rec} & \textbf{F1} \\
\midrule

\multirow{9}{*}{\rotatebox{90}{\textbf{Iris}}}
& \cellcolor{gray!10}Centralized SVDD & \multicolumn{5}{c|}{\cellcolor{gray!10}\textit{-- Centralized Baselines --}} & \multicolumn{5}{c}{\cellcolor{gray!10}\textit{-- Centralized Baselines --}} \\
& \cellcolor{gray!10}Centralized k-NN & \multicolumn{5}{c|}{\cellcolor{gray!10}\textit{}} & \multicolumn{5}{c}{\cellcolor{gray!10}\textit{}} \\
& FedAvg (FE) & 90.70$\pm$9.50 & 82.40$\pm$13.3 & 91.90$\pm$7.50 & 82.20$\pm$12.9 & 85.00$\pm$12.4 & 90.10$\pm$10.4 & 82.90$\pm$13.4 & 94.10$\pm$5.70 & 81.60$\pm$13.5 & 85.20$\pm$12.5 \\
& FedAvg (No SM) & \textbf{98.32$\pm$1.64} & \textbf{92.89$\pm$4.69} & 96.27$\pm$3.11 & \textbf{92.92$\pm$4.57} & \textbf{94.34$\pm$3.92} & \textbf{97.36$\pm$2.91} & \underline{90.89$\pm$7.41} & 93.93$\pm$5.77 & \textbf{92.53$\pm$5.80} & \underline{92.77$\pm$6.16} \\
& FedAvg (SM) & 95.81$\pm$3.29 & 87.78$\pm$7.25 & \underline{96.56$\pm$2.47} & 85.15$\pm$8.32 & 89.83$\pm$5.93 & 96.30$\pm$2.98 & 86.67$\pm$8.02 & \underline{96.52$\pm$2.64} & 83.45$\pm$9.37 & 88.78$\pm$6.68 \\
& Fed-Forest & \underline{96.16$\pm$4.20} & \underline{90.44$\pm$6.76} & \textbf{96.81$\pm$4.42} & \underline{89.07$\pm$9.54} & \underline{92.43$\pm$5.60} & \underline{96.49$\pm$3.83} & \textbf{92.22$\pm$4.66} & \textbf{97.13$\pm$3.48} & \underline{91.26$\pm$6.49} & \textbf{93.94$\pm$3.61} \\
& Fed-XGB & 80.41$\pm$13.0 & 74.67$\pm$13.3 & 87.44$\pm$9.98 & 74.09$\pm$19.2 & 78.29$\pm$13.8 & 82.51$\pm$14.2 & 74.89$\pm$15.5 & 88.76$\pm$9.59 & 72.01$\pm$23.0 & 77.36$\pm$16.2 \\
\cmidrule{2-12}
\rowcolor{geoPrimary!10} & \textbf{Geodesic (OS)} & 92.10$\pm$6.90 & 85.10$\pm$10.9 & 96.20$\pm$3.90 & 80.40$\pm$14.3 & 86.80$\pm$10.5 & 92.90$\pm$5.90 & 86.00$\pm$9.60 & 93.30$\pm$7.00 & 85.10$\pm$9.20 & 88.60$\pm$8.00 \\
\rowcolor{geoPrimary!10} & \textbf{Geodesic (Iter)} & 89.80$\pm$10.1 & 82.90$\pm$13.7 & 94.00$\pm$5.80 & 81.00$\pm$14.5 & 85.10$\pm$12.9 & 90.70$\pm$9.30 & 83.10$\pm$12.5 & 94.00$\pm$4.60 & 81.40$\pm$14.5 & 85.10$\pm$12.3 \\
\midrule

\multirow{9}{*}{\rotatebox{90}{\textbf{BreastC}}}
& \cellcolor{gray!10}Centralized SVDD & \multicolumn{5}{c|}{\cellcolor{gray!10}\textit{-- Centralized Baselines --}} & \multicolumn{5}{c}{\cellcolor{gray!10}\textit{-- Centralized Baselines --}} \\
& \cellcolor{gray!10}Centralized k-NN & \multicolumn{5}{c|}{\cellcolor{gray!10}\textit{}} & \multicolumn{5}{c}{\cellcolor{gray!10}\textit{}} \\
& FedAvg (FE) & \textbf{99.20$\pm$0.10} & \textbf{96.60$\pm$0.40} & \textbf{97.40$\pm$0.70} & \underline{95.20$\pm$2.30} & \textbf{96.20$\pm$0.90} & \textbf{99.10$\pm$0.00} & \textbf{96.30$\pm$0.40} & \textbf{96.70$\pm$0.20} & \textbf{95.20$\pm$1.80} & \textbf{95.80$\pm$1.10} \\
& FedAvg (No SM) & 98.48$\pm$0.27 & 93.51$\pm$1.75 & 94.40$\pm$1.26 & 92.37$\pm$0.59 & 93.23$\pm$0.42 & 98.48$\pm$0.38 & 93.33$\pm$2.63 & 94.82$\pm$0.33 & 92.09$\pm$1.32 & 93.30$\pm$0.46 \\
& FedAvg (SM) & 98.77$\pm$0.32 & 94.56$\pm$1.05 & 95.13$\pm$1.66 & 93.76$\pm$0.29 & 94.29$\pm$0.79 & 98.45$\pm$0.43 & 94.21$\pm$1.75 & 95.69$\pm$0.32 & 92.83$\pm$0.13 & 94.08$\pm$0.21 \\
& Fed-Forest & 91.58$\pm$6.21 & 84.04$\pm$7.07 & 86.55$\pm$7.64 & 81.09$\pm$8.99 & 83.15$\pm$5.71 & 92.61$\pm$5.04 & 84.39$\pm$6.16 & 87.59$\pm$8.42 & 80.91$\pm$5.65 & 83.68$\pm$4.36 \\
& Fed-XGB & 70.06$\pm$19.0 & 69.39$\pm$18.1 & 68.62$\pm$10.8 & 71.20$\pm$30.7 & 66.00$\pm$24.7 & 63.65$\pm$24.2 & 62.72$\pm$20.7 & 64.22$\pm$13.7 & 57.66$\pm$33.6 & 55.46$\pm$27.8 \\
\cmidrule{2-12}
\rowcolor{geoPrimary!10} & \textbf{Geodesic (OS)} & 98.60$\pm$0.00 & 93.80$\pm$0.30 & 92.60$\pm$3.00 & 93.40$\pm$2.20 & 92.90$\pm$2.60 & 97.80$\pm$0.10 & 92.50$\pm$0.20 & 90.10$\pm$4.30 & 93.30$\pm$1.20 & 91.60$\pm$2.80 \\
\rowcolor{geoPrimary!10} & \textbf{Geodesic (Iter)} & \underline{99.10$\pm$0.00} & \underline{96.30$\pm$0.50} & \underline{96.70$\pm$0.20} & \textbf{95.40$\pm$1.30} & \underline{96.00$\pm$0.80} & \underline{99.00$\pm$0.10} & \underline{96.00$\pm$0.50} & \underline{96.50$\pm$0.60} & \underline{94.80$\pm$1.20} & \underline{95.60$\pm$0.90} \\
\midrule

\multirow{9}{*}{\rotatebox{90}{\textbf{Titanic}}}
& \cellcolor{gray!10}Centralized SVDD & \multicolumn{5}{c|}{\cellcolor{gray!10}\textit{-- Centralized Baselines --}} & \multicolumn{5}{c}{\cellcolor{gray!10}\textit{-- Centralized Baselines --}} \\
& \cellcolor{gray!10}Centralized k-NN & \multicolumn{5}{c|}{\cellcolor{gray!10}\textit{}} & \multicolumn{5}{c}{\cellcolor{gray!10}\textit{}} \\
& FedAvg (FE) & \underline{71.90$\pm$0.20} & \underline{69.20$\pm$0.40} & 68.90$\pm$10.3 & \underline{69.90$\pm$3.30} & \underline{68.10$\pm$3.30} & \underline{73.80$\pm$0.30} & \underline{70.10$\pm$0.20} & \textbf{70.10$\pm$9.00} & 68.80$\pm$1.00 & \underline{68.50$\pm$3.50} \\
& FedAvg (No SM) & 64.88$\pm$10.3 & 62.25$\pm$9.12 & \underline{70.09$\pm$5.38} & 47.38$\pm$14.4 & 54.42$\pm$8.76 & 64.87$\pm$9.81 & 61.68$\pm$9.31 & 67.98$\pm$3.84 & 48.00$\pm$13.8 & 54.38$\pm$8.55 \\
& FedAvg (SM) & 63.49$\pm$0.97 & 61.98$\pm$0.31 & 59.34$\pm$8.50 & 67.45$\pm$1.74 & 62.04$\pm$6.08 & 64.66$\pm$0.04 & 63.93$\pm$1.87 & 60.40$\pm$9.78 & 69.29$\pm$5.66 & 63.54$\pm$8.40 \\
& Fed-Forest & 63.03$\pm$12.1 & 60.11$\pm$9.88 & 63.55$\pm$8.31 & 52.11$\pm$22.0 & 53.62$\pm$15.3 & 63.75$\pm$12.5 & 57.82$\pm$11.2 & 65.83$\pm$10.7 & 40.49$\pm$19.4 & 46.26$\pm$15.1 \\
& Fed-XGB & 56.83$\pm$12.4 & 55.76$\pm$11.8 & 62.53$\pm$8.62 & 37.86$\pm$22.7 & 41.98$\pm$18.9 & 57.33$\pm$7.36 & 58.66$\pm$7.78 & 63.17$\pm$8.05 & 46.47$\pm$17.5 & 50.70$\pm$12.3 \\
\cmidrule{2-12}
\rowcolor{geoPrimary!10} & \textbf{Geodesic (OS)} & 61.20$\pm$8.60 & 61.50$\pm$5.00 & 57.10$\pm$18.5 & 66.50$\pm$6.70 & 58.20$\pm$13.7 & 59.00$\pm$7.20 & 59.20$\pm$6.00 & 59.60$\pm$12.8 & \textbf{75.90$\pm$0.90} & 63.50$\pm$8.00 \\
\rowcolor{geoPrimary!10} & \textbf{Geodesic (Iter)} & \textbf{75.30$\pm$1.60} & \textbf{72.50$\pm$2.10} & \textbf{70.30$\pm$9.50} & \textbf{74.50$\pm$2.90} & \textbf{72.10$\pm$6.40} & \textbf{75.30$\pm$1.50} & \textbf{72.60$\pm$1.90} & \underline{70.00$\pm$8.70} & \underline{74.80$\pm$3.90} & \textbf{72.20$\pm$6.50} \\
\midrule

\multirow{9}{*}{\rotatebox{90}{\textbf{Adult}}}
& \cellcolor{gray!10}Centralized SVDD & \multicolumn{5}{c|}{\cellcolor{gray!10}\textit{-- Centralized Baselines --}} & \multicolumn{5}{c}{\cellcolor{gray!10}\textit{-- Centralized Baselines --}} \\
& \cellcolor{gray!10}Centralized k-NN & \multicolumn{5}{c|}{\cellcolor{gray!10}\textit{}} & \multicolumn{5}{c}{\cellcolor{gray!10}\textit{}} \\
& FedAvg (FE) & 61.50$\pm$18.9 & 65.90$\pm$4.00 & 60.00$\pm$32.3 & 48.60$\pm$17.1 & 50.00$\pm$26.8 & 61.90$\pm$18.5 & 69.10$\pm$0.10 & 60.10$\pm$32.4 & 40.60$\pm$23.7 & 48.30$\pm$27.6 \\
& FedAvg (No SM) & 62.36$\pm$18.3 & \textbf{70.15$\pm$0.52} & 60.15$\pm$31.9 & 41.48$\pm$25.5 & 48.88$\pm$28.6 & \underline{62.43$\pm$18.4} & \underline{69.78$\pm$0.72} & 60.69$\pm$32.2 & 38.95$\pm$25.0 & 47.15$\pm$28.6 \\
& FedAvg (SM) & 59.78$\pm$1.31 & 58.25$\pm$9.59 & \underline{62.90$\pm$26.4} & 43.41$\pm$6.02 & 45.06$\pm$6.04 & 58.39$\pm$3.06 & 51.28$\pm$2.85 & \underline{61.17$\pm$30.0} & 50.84$\pm$14.6 & 45.72$\pm$4.17 \\
& Fed-Forest & \textbf{65.34$\pm$7.62} & 69.44$\pm$3.00 & \textbf{63.10$\pm$24.0} & \underline{51.69$\pm$17.7} & \textbf{56.44$\pm$20.4} & \textbf{65.54$\pm$7.14} & \textbf{71.84$\pm$2.38} & \textbf{65.15$\pm$21.1} & \underline{51.02$\pm$21.2} & \underline{57.00$\pm$21.4} \\
& Fed-XGB & 43.08$\pm$11.9 & 57.96$\pm$20.2 & 61.31$\pm$21.0 & 15.04$\pm$20.2 & 20.77$\pm$23.2 & 36.69$\pm$7.34 & 47.84$\pm$23.1 & 59.53$\pm$22.3 & 15.67$\pm$29.1 & 13.23$\pm$16.7 \\
\cmidrule{2-12}
\rowcolor{geoPrimary!10} & \textbf{Geodesic (OS)} & \underline{62.40$\pm$9.30} & 68.00$\pm$0.30 & 61.20$\pm$25.3 & \textbf{61.50$\pm$10.8} & \underline{53.30$\pm$22.5} & 62.20$\pm$4.90 & 62.60$\pm$6.90 & 59.40$\pm$25.0 & \textbf{70.20$\pm$4.10} & \textbf{59.50$\pm$18.6} \\
\rowcolor{geoPrimary!10} & \textbf{Geodesic (Iter)} & 61.60$\pm$18.8 & \underline{69.60$\pm$0.20} & 59.60$\pm$32.6 & 43.10$\pm$22.8 & 49.80$\pm$27.0 & 61.20$\pm$19.0 & 69.20$\pm$0.50 & 59.70$\pm$32.8 & 40.90$\pm$24.3 & 48.40$\pm$28.1 \\
\midrule

\multirow{9}{*}{\rotatebox{90}{\textbf{Covertype}}}
& \cellcolor{gray!10}Centralized SVDD & \multicolumn{5}{c|}{\cellcolor{gray!10}\textit{-- Centralized Baselines --}} & \multicolumn{5}{c}{\cellcolor{gray!10}\textit{-- Centralized Baselines --}} \\
& \cellcolor{gray!10}Centralized k-NN & \multicolumn{5}{c|}{\cellcolor{gray!10}\textit{}} & \multicolumn{5}{c}{\cellcolor{gray!10}\textit{}} \\
& FedAvg (FE) & 85.00$\pm$18.7 & 80.20$\pm$17.1 & \textbf{95.80$\pm$6.70} & 73.60$\pm$27.1 & 78.30$\pm$26.6 & 86.90$\pm$17.7 & 82.10$\pm$16.7 & 94.40$\pm$9.10 & 75.30$\pm$27.3 & 81.10$\pm$24.0 \\
& FedAvg (No SM) & \textbf{91.62$\pm$9.83} & \textbf{85.39$\pm$11.2} & \underline{94.88$\pm$8.46} & \textbf{81.93$\pm$16.3} & \textbf{87.54$\pm$13.2} & \textbf{91.47$\pm$10.1} & \textbf{85.33$\pm$11.3} & \textbf{95.00$\pm$8.11} & \textbf{81.65$\pm$16.7} & \textbf{87.40$\pm$13.4} \\
& FedAvg (SM) & 77.29$\pm$14.5 & 67.03$\pm$16.3 & 92.43$\pm$13.1 & 61.46$\pm$21.3 & 71.84$\pm$18.7 & 76.86$\pm$13.9 & 65.69$\pm$17.3 & 93.80$\pm$10.6 & 58.38$\pm$23.2 & 69.62$\pm$19.9 \\
& Fed-Forest & 70.70$\pm$8.46 & 69.66$\pm$10.8 & 88.75$\pm$15.5 & 69.28$\pm$12.8 & 77.43$\pm$13.2 & 70.82$\pm$8.77 & 71.52$\pm$11.7 & 88.71$\pm$15.6 & 70.59$\pm$15.2 & 78.22$\pm$15.1 \\
& Fed-XGB & 73.83$\pm$13.2 & 64.75$\pm$11.5 & 89.34$\pm$15.9 & 61.09$\pm$16.1 & 71.96$\pm$15.8 & 72.46$\pm$15.2 & 61.28$\pm$12.9 & 88.63$\pm$17.4 & 56.86$\pm$21.1 & 67.57$\pm$19.7 \\
\cmidrule{2-12}
\rowcolor{geoPrimary!10} & \textbf{Geodesic (OS)} & 81.70$\pm$13.0 & 74.30$\pm$12.7 & 91.30$\pm$14.0 & 72.50$\pm$14.4 & 79.70$\pm$14.4 & 76.00$\pm$15.9 & 69.50$\pm$15.4 & 89.10$\pm$16.8 & 70.60$\pm$14.6 & 75.70$\pm$16.4 \\
\rowcolor{geoPrimary!10} & \textbf{Geodesic (Iter)} & \underline{91.20$\pm$9.60} & \underline{84.60$\pm$11.0} & 94.50$\pm$9.00 & \underline{81.50$\pm$15.2} & \underline{87.30$\pm$12.7} & \underline{90.70$\pm$10.1} & \underline{84.30$\pm$11.2} & \underline{94.80$\pm$8.40} & \underline{80.30$\pm$16.6} & \underline{86.50$\pm$13.5} \\
\bottomrule
\end{tabular}
}
\end{table*}

\begin{table}[t]
    \centering
    \caption{\textbf{Comparison of Linear (LERP) vs. Geodesic (SLERP)}. $\uparrow$ Higher is better, $\downarrow$ Lower is better.}
    \label{tab:ablation_study}
    \resizebox{\columnwidth}{!}{%
    \begin{tabular}{l|cc|c}
        \toprule
        \multirow{2}{*}{\textbf{Metric}} & \multicolumn{2}{c|}{\textbf{Interpolation Method}} & \multirow{2}{*}{\textbf{Status}} \\
        & \textbf{Linear (LERP)} & \textbf{Geodesic (SLERP)} & \\
        \midrule
        \rowcolor{geoPrimary!10}  \multicolumn{4}{l}{\textit{Downstream Performance (Covertype, One-Shot)}} \\
        AUROC Score ($\uparrow$) & $0.920 \pm 0.08$ & $\mathbf{0.921 \pm 0.08}$ & \textit{Parity} \\
        Accuracy ($\uparrow$) & $0.851 \pm 0.10$ & $\mathbf{0.856 \pm 0.10}$ & \textit{Parity} \\
        F1-Score ($\uparrow$) & $0.880 \pm 0.11$ & $\mathbf{0.882 \pm 0.12}$ & \textit{Parity} \\
        \midrule
        \rowcolor{geoPrimary!10} \multicolumn{4}{l}{\textit{Geometric Integrity \& Stability}} \\
        \textbf{Constraint ($\|z\| \equiv 1$)} & \textcolor{red}{\textbf{Violated}} & \textcolor{teal}{\textbf{Satisfied}} & \textit{Validity} \\
        Gradient Var. ($\downarrow$) & $3.99 \times 10^{-3}$ & $\mathbf{5.50 \times 10^{-12}}$ & \textit{Stable} \\
       Midpoint Norm, random pairs ($\uparrow$) & $0.705 \pm 0.03$ & $\mathbf{1.000 \pm 0.00}$ & \textit{Adherence} \\
        Loss Dev. ($\downarrow$) & $0.078$ & $\mathbf{0.000}$ & \textit{Exactness} \\
        \bottomrule
    \end{tabular}%
    }
\end{table}

\section{Conclusion}
\label{sec:conclusion}

In this work, we addressed the fundamental incompatibility between state-of-the-art tabular learning and Federated Learning—the “Federated Tabular Gap.” We introduced \textit{Geodesic-SVDD}, a framework for \textit{Distributed Spherical Prototype Learning} that recasts federated tabular learning as one-class density estimation on the hypersphere. Our results indicate that this recasting is what neutralizes label skew: under a single-center objective, iterative weight averaging remains stable even when clients hold nearly disjoint classes, with Geodesic-SVDD (Iter) staying within $1.5$ AUROC points of its IID performance on every benchmark. By mapping each client's data to a compact prototype on a hyperspherical manifold ($\mathbb{S}^{d-1}$), the same formulation enables a \textit{Product of Experts (PoE)} aggregation protocol that requires a single upload per client.
In the One-Shot setting, this additive density estimation lets the global model form a ``Union of Manifolds'' from a single upload per client, matching centralized Deep SVDD under IID and mild skew. With iterative alignment, Geodesic-SVDD attains the highest mean AUROC among the seven federated methods we evaluate and outperforms federated tree ensembles by up to 21 AUROC points on \textit{Covertype}.


Our results suggest a clear decision rule for practitioners based on the ``Privacy-Utility-Skew'' triangle: \textit{High-Security / Low-Bandwidth (One-Shot):} In scenarios where iterative communication is prohibitive (e.g., cross-silo financial fraud detection), our One-Shot protocol offers a highly efficient solution. By uploading a single, valid probabilistic manifold, clients match centralized Deep SVDD on IID and mild-skew data without ever exposing gradient updates. \textit{Extreme Heterogeneity (Iterative):} In ``worst-case'' regimes where clients hold mutually exclusive classes (Extreme Skew, $\alpha=0.01$), the iterative alignment of these manifolds becomes necessary, recovering the One-Shot losses (e.g., from $0.76$ to $0.91$ AUROC on \textit{Covertype}).

While our focus is on the architectural framework, our ablation study highlights a critical implementation detail. Although linear approximations (Linear-SVDD) can empirically match the predictive performance of rigorous geodesic interpolation, they do so at the cost of numerical stability ($10^9\times$ gradient variance). For safety-critical systems, strictly enforcing the Riemannian geometry via SLERP is required to guarantee reliable convergence, ensuring that the theoretical ``normality'' constraints of the SVDD objective are never violated.

Finally, future work will explore extending this spherical prototype approach to encompass \textit{Riemannian Differential Privacy}~\cite{riemannian_dp_neurips}. Standard DP mechanisms require artificial gradient clipping to bound sensitivity, often introducing bias that degrades utility. In contrast, our latent space is naturally compact ($\|\phi(x)\| \equiv 1$), meaning the sensitivity of any query is strictly bounded by the geodesic diameter of the manifold ($\pi$)~\cite{fed_riemannian_dp}. We hypothesize that leveraging this intrinsic geometric bound—coupled with Riemannian Gaussian noise injection—will yield a superior privacy-utility trade-off compared to extrinsic Euclidean clipping methods.

\bibliographystyle{IEEEtran}
\bibliography{Reference}
\end{document}

%% file: diagram.tex
\begin{tikzpicture}

    
    \node (TopTL) at (0, 2.8) {}; 
    \node (TopBR) at (24, -2.4) {};
    
    \node (BotTL) at (0, -2.8) {};
    \node (BotBR) at (24, -7.6) {};

    
    \node[font=\bfseries\sffamily\small, text=geoDark, align=center] (InputLabel) at (1.5, 1.8) {Tabular Input\\$x \in \mathbb{R}^D$};
    
    \begin{scope}[local bounding box=InputVector, shift={(1.5, 0)}]
        \node[dataNode] at (0, 0.7) {\scalebox{0.75}{$x_1$}};
        \node[dataNode] at (0, 0.35) {\scalebox{0.75}{$x_2$}};
        \node[dataNode] at (0, 0.0) {\scalebox{0.75}{$\vdots$}};
        \node[dataNode] at (0, -0.35) {\scalebox{0.75}{$x_{\scriptscriptstyle D-1}$}};
        \node[dataNode] at (0, -0.7) {\scalebox{0.75}{$x_{\scriptscriptstyle D}$}};
    \end{scope}

    \node[layerDense, font=\small] (Proj) at (3.5, 0) {$W_{\text{in}}$};
    \node[above=0.05cm of Proj, font=\tiny\sffamily] {Proj};
    
    \draw[flow] (InputVector.east) -- node[above, font=\tiny, yshift=1pt] {$\mathbf{x}$} (Proj.west);

    \node[layerInternal, font=\scriptsize] (L1) at (5.5, 0) {$W_1$}; 
    \node[opCircle, scale=0.7] (Act1) at (6.2, 0) {$\sigma$};
    \node[layerInternal, font=\scriptsize] (L2) at (6.9, 0) {$W_2$};
    \node[opCircle, scale=0.7] (Act2) at (7.6, 0) {$\sigma$};
    \node[opCircle, scale=0.8] (Add) at (8.5, 0) {+};
    
    \draw[flow, thin] (L1.east) -- (Act1.west);
    \draw[flow, thin] (Act1.east) -- (L2.west);
    \draw[flow, thin] (L2.east) -- (Act2.west);
    \draw[flow, thin] (Act2.east) -- (Add.west);

    \begin{scope}[on background layer]
        \node[layerResContainer, fit=(L1)(Add), inner sep=7pt] (ResBlock) {};
    \end{scope}
    \node[above=0.05cm of ResBlock, font=\bfseries\sffamily\scriptsize, text=geoDark, align=center] {Residual Encoder\\ \tiny $h \leftarrow h + \mathcal{F}(h)$};

    \draw[flow] (Proj.east) -- node[above, font=\tiny, pos=0.4, yshift=1pt] {$h_0$} (L1.west);
    \draw[residFlow] ($(L1.west)+(-0.2,0)$) -- ++(0, 1.0) -| (Add.north);

    \node[layerDense, font=\small] (Head) at (10, 0) {$W_{\text{head}}$};
    \node[above=0.05cm of Head, font=\tiny\sffamily] {Head};
    \draw[flow] (Add.east) -- node[above, font=\tiny, yshift=1pt] {$h_L$} (Head.west);

    \node[normBlock] (Norm) at (12.6, 0) {Normalize\\$z = \frac{\hat{z}}{\|\hat{z}\|_2}$};
    \draw[flow] (Head.east) -- node[above, font=\tiny, yshift=1pt] {$\hat{z}$} (Norm.west);

    \begin{scope}[shift={(16.5,0)}] 
        \fill[geoPrimary!5] (0,0) circle (1.6cm); 
        \shade[ball color=geoPrimary!20, opacity=0.4] (0,0) circle (1.6cm); 

        \draw[geoPrimary!30, thick] (0,0) circle (1.6cm);
        \draw[geoPrimary!20, dashed] (-1.6,0) arc (180:360:1.6cm and 0.4cm);
        \draw[geoPrimary!20] (1.6,0) arc (0:180:1.6cm and 0.4cm);
        \draw[geoPrimary!20] (0,1.6) arc (90:270:0.4cm and 1.6cm);

        \coordinate (O) at (0,0);
        \fill[geoDark] (O) circle (1.5pt);
        \coordinate (Zi) at (110:1.6cm);
        \draw[vec, geoDark] (O) -- (Zi) node[above left, font=\scriptsize, inner sep=1pt] {$z_i$};
        \fill[geoDark] (Zi) circle (2pt);
        \coordinate (Zj) at (40:1.6cm);
        \draw[vec, geoDark] (O) -- (Zj) node[right, font=\scriptsize, inner sep=1pt] {$z_j$};
        \fill[geoDark] (Zj) circle (2pt);
        \draw[geoAccent, line width=1.5pt] (Zj) arc (40:110:1.6cm);
        \coordinate (Zmix) at (75:1.6cm);
        \fill[geoAccent] (Zmix) circle (2.5pt);
        
        \node[below=0.08cm of Zmix, text=geoAccent, font=\bfseries\tiny] {$\tilde{z}_{geo}$};
        
        \node[font=\bfseries\scriptsize, text=geoPrimary] at (0, 2.0) {Spherical Interpolation};
        \coordinate (SphereLeft) at (-1.8,0);
    \end{scope}

    \begin{scope}[shift={(21,0)}]
        \fill[linearColor!5] (0,0) circle (1.6cm); 
        \shade[ball color=linearColor!20, opacity=0.4] (0,0) circle (1.6cm); 

        \draw[linearColor!40, thick] (0,0) circle (1.6cm); 
        \draw[linearColor!20, dashed] (-1.6,0) arc (180:360:1.6cm and 0.4cm);
        \draw[linearColor!20] (1.6,0) arc (0:180:1.6cm and 0.4cm);
        \draw[linearColor!20] (0,1.6) arc (90:270:0.4cm and 1.6cm);

        \coordinate (O2) at (0,0);
        \fill[geoDark] (O2) circle (1.5pt);
        \coordinate (Zi2) at (110:1.6cm);
        \draw[vec, geoDark] (O2) -- (Zi2) node[above left, font=\scriptsize, inner sep=1pt] {$z_i$};
        \fill[geoDark] (Zi2) circle (2pt);
        \coordinate (Zj2) at (40:1.6cm);
        \draw[vec, geoDark] (O2) -- (Zj2) node[right, font=\scriptsize, inner sep=1pt] {$z_j$};
        \fill[geoDark] (Zj2) circle (2pt);
        \draw[linearColor, line width=1.2pt] (Zi2) -- (Zj2);
        \coordinate (ZmixLin) at ($(Zi2)!0.5!(Zj2)$);
        \fill[linearColor] (ZmixLin) circle (2.2pt);
        \node[below=0.05cm of ZmixLin, text=linearColor, font=\bfseries\tiny] {$\tilde{z}_{lin}$};
        \node[font=\bfseries\scriptsize, text=linearColor] at (0, 2.0) {Linear Interpolation};
    \end{scope}

    \draw[vec, dashed, geoDark] (Norm.east) -- node[above, font=\tiny, yshift=2pt] {$\phi(x)$} (SphereLeft);
    \draw[dashed, gray!25, thick] (18.75, 1.6) -- (18.75, -1.6);

    
    \node[anchor=center, font=\bfseries\scriptsize, text=geoPrimary] at (6, -3.3) {One-Shot Federated (OSFL)};
    \node[anchor=center, font=\bfseries\scriptsize, text=linearColor] at (18, -3.3) {Iterative Federated (FL)};
    
    \node[clientBox, label={left:\tiny\color{gray}C1}] (C1_OS) at (3.5, -4.2) {
        Train $\theta_1$\\
        \tiny $\min \mathbb{E}[\ell_{geo}]$
    };
    \node[clientBox, label={left:\tiny\color{gray}C2}] (C2_OS) at (3.5, -5.3) {
        Train $\theta_2$\\
        \tiny $\min \mathbb{E}[\ell_{geo}]$
    };
    \node[clientBox, label={left:\tiny\color{gray}C3}] (C3_OS) at (3.5, -6.4) {
        Train $\theta_3$\\
        \tiny $\min \mathbb{E}[\ell_{geo}]$
    };
    
    \node[serverBox, fill=geoPrimary!10, draw=geoPrimary] (Server_OS) at (8.5, -5.3) {
        \textbf{Global Server}\\
        \scriptsize Product of Experts (PoE)\\
        \tiny $s(x) = -\sum_{k} \log p(y|\phi_k)$
    };
    
    \draw[flow, geoPrimary] (C1_OS.east) -- (Server_OS.west|-C1_OS.east);
    \draw[flow, geoPrimary] (C2_OS.east) -- (Server_OS.west);
    \draw[flow, geoPrimary] (C3_OS.east) -- (Server_OS.west|-C3_OS.east);

    \draw[dotted, gray!35, thick] (12.6, -3.5) -- (12.6, -6.9);
    
    \node[clientBox, label={left:\tiny\color{gray}C1}] (C1_FL) at (16.5, -4.2) {
        Update $\mathcal{W}_1^t$\\
        \tiny $\mathcal{W} - \eta \nabla \mathcal{L}_1$
    };
    \node[clientBox, label={left:\tiny\color{gray}C2}] (C2_FL) at (16.5, -5.3) {
        Update $\mathcal{W}_2^t$\\
        \tiny $\mathcal{W} - \eta \nabla \mathcal{L}_2$
    };
    \node[clientBox, label={left:\tiny\color{gray}C3}] (C3_FL) at (16.5, -6.4) {
        Update $\mathcal{W}_3^t$\\
        \tiny $\mathcal{W} - \eta \nabla \mathcal{L}_3$
    };
    
    \node[serverBox] (Server_FL) at (20.5, -5.3) {
        \textbf{Global Server}\\
        \scriptsize FedAvg Aggregation\\
        \tiny $\mathcal{W}^{t+1} \leftarrow \sum_{k} \frac{N_k}{N} \mathcal{W}_k^{t}$
    };
    
    \draw[flow, <->, linearColor] (C1_FL.east) -- (Server_FL.west|-C1_FL.east);
    \draw[flow, <->, linearColor] (C2_FL.east) -- (Server_FL.west);
    \draw[flow, <->, linearColor] (C3_FL.east) -- (Server_FL.west|-C3_FL.east);

    \begin{scope}[on background layer]
        
        \node[squircleStrip, draw=geoPrimary!30, fill=geoPrimary!3,
              fit=(TopTL)(TopBR)] (TopFrame) {};
        
        \node[stripLabel, text=geoPrimary] at (TopFrame.west) {Local Client Processing};

        \node[squircleStrip, draw=linearColor!30, fill=linearColor!3,
              fit=(BotTL)(BotBR)] (BottomFrame) {};
        
        \node[stripLabel, text=linearColor] at (BottomFrame.west) {Server Aggregation Protocols};

    \end{scope}

\end{tikzpicture}

%% file: Reference.bib
@inproceedings{zhang2017mixup,
  title     = {mixup: Beyond Empirical Risk Minimization},
  author    = {Zhang, Hongyi and Cisse, Moustapha and Dauphin, Yann N. and Lopez-Paz, David},
  booktitle = {International Conference on Learning Representations (ICLR)},
  year      = {2018}
}

@inproceedings{riemannian_dp_neurips,
  title     = {Differential Privacy Over {R}iemannian Manifolds},
  author    = {Reimherr, Matthew and Bharath, Karthik and Soto, Carlos},
  booktitle = {Advances in Neural Information Processing Systems (NeurIPS)},
  volume    = {34},
  year      = {2021}
}

@inproceedings{verma2019manifold,
  title     = {Manifold Mixup: Better Representations by Interpolating Hidden States},
  author    = {Verma, Vikas and Lamb, Alex and Beckham, Christopher and Najafi, Amir and Mitliagkas, Ioannis and Lopez-Paz, David and Bengio, Yoshua},
  booktitle = {Proceedings of the 36th International Conference on Machine Learning (ICML)},
  series    = {PMLR},
  volume    = {97},
  pages     = {6438--6447},
  year      = {2019}
}

@inproceedings{chen2016xgboost,
  title     = {{XGBoost}: A Scalable Tree Boosting System},
  author    = {Chen, Tianqi and Guestrin, Carlos},
  booktitle = {Proceedings of the 22nd ACM SIGKDD International Conference on Knowledge Discovery and Data Mining},
  pages     = {785--794},
  year      = {2016}
}

@article{benavoli2017time,
  author  = {Alessio Benavoli and Giorgio Corani and Janez Dem{\v{s}}ar and Marco Zaffalon},
  title   = {Time for a Change: a Tutorial for Comparing Multiple Classifiers Through Bayesian Analysis},
  journal = {Journal of Machine Learning Research},
  year    = {2017},
  volume  = {18},
  number  = {77},
  pages   = {1--36},
  url     = {http://jmlr.org/papers/v18/16-305.html}
}

@misc{covertype_31,
  author       = {Blackard, Jock},
  title        = {{Covertype}},
  year         = {1998},
  howpublished = {UCI Machine Learning Repository},
  note         = {{DOI}: https://doi.org/10.24432/C50K5N}
}

@article{JMLRdemsar06a,
  author  = {Janez Dem{\v{s}}ar},
  title   = {Statistical Comparisons of Classifiers over Multiple Data Sets},
  journal = {Journal of Machine Learning Research},
  year    = {2006},
  volume  = {7},
  number  = {1},
  pages   = {1--30}
}

@article{zhao2018federated,
  title={Federated learning with non-iid data},
  author={Zhao, Yue and Li, Meng and Lai, Liangzhen and Suda, Naveen and Civin, Damon and Chandra, Vikas},
  journal={arXiv preprint arXiv:1806.00582},
  year={2018},
  url={https://arxiv.org/abs/1806.00582}
}

@article{li2020federated,
  title={Federated learning: Challenges, methods, and future directions},
  author={Li, Tian and Sahu, Anit Kumar and Talwalkar, Ameet and Smith, Virginia},
  journal={IEEE signal processing magazine},
  volume={37},
  number={3},
  pages={50--60},
  year={2020},
  publisher={IEEE}
}

@article{ruopp2008youden,
  title={Youden Index and optimal cut-point estimated from observations affected by a lower limit of detection},
  author={Ruopp, MD and Perkins, NJ and Whitcomb, BW and Schisterman, EF},
  journal={Biometrical Journal},
  volume={50},
  number={3},
  year={2008},
  publisher={Wiley Online Library},
  doi={10.1002/bimj.200710415},
  pmid={18435502},
  pmcid={PMC2515362}
}

@inproceedings{ruff2018deep,
  title={Deep one-class classification},
  author={Ruff, Lukas and Vandermeulen, Robert and Goernitz, Nico and Deecke, Lucas and Siddiqui, Shoaib Ahmed and Binder, Alexander and M{\"u}ller, Emmanuel and Kloft, Marius},
  booktitle={International Conference on Machine Learning},
  year={2018},
  organization={PMLR}
}

@article{cover1967nearest,
  title={Nearest neighbor pattern classification},
  author={Cover, Thomas and Hart, Peter},
  journal={IEEE Transactions on Information Theory},
  volume={13},
  number={1},
  year={1967},
  publisher={IEEE}
}

@inproceedings{mcmahan2017communication,
  title={Communication-efficient learning of deep networks from decentralized data},
  author={McMahan, Brendan and Moore, Eider and Ramage, Daniel and Hampson, Seth and Arcas, Blaise Aguera y},
  booktitle={Artificial Intelligence and Statistics},
  year={2017},
  organization={PMLR}
}

@INPROCEEDINGS{liu2020federated,
  author={Liu, Fei Tony and Ting, Kai Ming and Zhou, Zhi-Hua},
  booktitle={2008 Eighth IEEE International Conference on Data Mining}, 
  title={Isolation Forest}, 
  year={2008},
  volume={},
  number={},
  pages={413-422},
  doi={10.1109/ICDM.2008.17}}

@article{cheng2021secureboost,
  title={Secureboost: A lossless federated learning framework},
  author={Cheng, Kewei and Fan, Tao and Jin, Yilun and Liu, Yang and Chen, Tianjian and Papadopoulos, Dimitrios and Yang, Qiang},
  journal={IEEE Intelligent Systems},
  volume={36},
  number={6},
  pages={87--98},
  year={2021},
  publisher={IEEE}
}

@INPROCEEDINGS{10651455,
  author={Internò, Christian and Olhofer, Markus and Jin, Yaochu and Hammer, Barbara},
  booktitle={2024 International Joint Conference on Neural Networks (IJCNN)}, 
  title={Federated Loss Exploration for Improved Convergence on Non-IID Data}, 
  year={2024},
  volume={},
  number={},
  pages={1-8},
  doi={10.1109/IJCNN60899.2024.10651455}}

@inproceedings{grinsztajn2022tree,
  title={Why do tree-based models still outperform deep learning on typical tabular data?},
  author={Grinsztajn, L{\'e}o and Oyallon, Edouard and Varoquaux, Ga{\"e}l},
  booktitle={NeurIPS Datasets and Benchmarks Track},
  year={2022}
}

@article{borisov2021deep,
   title={Deep Neural Networks and Tabular Data: A Survey},
   volume={35},
   ISSN={2162-2388},
   url={http://dx.doi.org/10.1109/TNNLS.2022.3229161},
   DOI={10.1109/tnnls.2022.3229161},
   number={6},
   journal={IEEE Transactions on Neural Networks and Learning Systems},
   publisher={Institute of Electrical and Electronics Engineers (IEEE)},
   author={Borisov, Vadim and Leemann, Tobias and Seßler, Kathrin and Haug, Johannes and Pawelczyk, Martin and Kasneci, Gjergji},
   year={2024},
   month=June, pages={7499–7519} }

@inproceedings{tan2022fedproto,
  title={FedProto: Federated Prototype Learning across Heterogeneous Clients},
  author={Tan, Yue and Long, Guodong and Liu, Lu and Zhou, Tianyi and Lu, Qinghua and Jiang, Jing and Zhang, Chengqi},
  booktitle={Proceedings of the AAAI Conference on Artificial Intelligence},
  volume={36},
  number={8},
  pages={8432--8440},
  year={2022},
  doi={10.1609/aaai.v36i8.20819}
}

@inproceedings{zhu2019deep,
  title={Deep leakage from gradients},
  author={Zhu, Ligeng and Liu, Zhijian and Han, Song},
  booktitle={Advances in neural information processing systems},
  volume={32},
  year={2019}
}

@inproceedings{guha2019one,
  title={One-shot federated learning},
  author={Guha, Neel and Talwalkar, Ameet and Smith, Virginia},
  booktitle={arXiv preprint arXiv:1902.11175},
  year={2019}
}

@ARTICLE{interno2024adaptive,
  author={Internò, Christian and Raponi, Elena and Olhofer, Markus and Raza, Ali and Bäck, Thomas and Stein, Niki van and Jin, Yaochu and Hammer, Barbara},
  journal={IEEE Internet of Things Journal}, 
  title={Pruning Federated Models Through Loss Landscape Analysis and Client Agreement Scoring}, 
  year={2026},
  volume={13},
  number={14},
  pages={31283-31293},
  doi={10.1109/JIOT.2026.3686028}}

@article{kairouz2021advances,
  title={Advances and open problems in federated learning},
  author={Kairouz, Peter and McMahan, H Brendan and Avent, Brendan and Bellet, Aur{\'e}lien and Bennis, Mehdi and Bhagoji, Arjun Nitin and Bonawitz, Kallista and Charles, Zachary and Cormode, Graham and Cummings, Rachel and others},
  journal={Foundations and Trends{\textregistered} in Machine Learning},
  volume={14},
  number={1--2},
  pages={1--210},
  year={2021},
  publisher={Now Publishers, Inc.}
}

@article{kirkpatrick2017overcoming,
  title={Overcoming catastrophic forgetting in neural networks},
  author={Kirkpatrick, James and Pascanu, Razvan and Rabinowitz, Neil and Veness, Joel and Desjardins, Guillaume and Rusu, Andrei A and Milan, Kieran and Quan, John and Ramalho, Tiago and Grabska-Barwinska, Agnieszka and others},
  journal={Proceedings of the national academy of sciences},
  volume={114},
  number={13},
  year={2017},
  publisher={National Acad Sciences}
}

@article{hinton2002training,
  title={Training products of experts by minimizing contrastive divergence},
  author={Hinton, Geoffrey E},
  journal={Neural computation},
  volume={14},
  number={8},
  pages={1771--1800},
  year={2002},
  publisher={MIT Press}
}

@inproceedings{shoemake1985animating,
  title={Animating rotation with quaternion curves},
  author={Shoemake, Ken},
  booktitle={Proceedings of the 12th annual conference on Computer graphics and interactive techniques},
  pages={245--254},
  year={1985}
}

@article{fed_riemannian_dp,
     title={Federated Learning on Riemannian Manifolds with Differential Privacy}, 
      author={Zhenwei Huang and Wen Huang and Pratik Jawanpuria and Bamdev Mishra},
      year={2024},
      eprint={2404.10029},
      archivePrefix={arXiv},
      primaryClass={math.OC},
      url={https://arxiv.org/abs/2404.10029}, 
}

@misc{iris53,
  author       = {Fisher, R. A.},
  title        = {{Iris}},
  year         = {1936},
  howpublished = {UCI Machine Learning Repository},
  note         = {{DOI}: https://doi.org/10.24432/C56C76}
}

@misc{kaggle_titanic,
  title = {Titanic - Machine Learning from Disaster},
  author = {Kaggle},
  year = {n.d.},
  url = {https://www.kaggle.com/competitions/titanic/data}
}

@misc{adult_2,
  author       = {Becker, Barry and Kohavi, Ronny},
  title        = {{Adult}},
  year         = {1996},
  howpublished = {UCI Machine Learning Repository},
  note         = {{DOI}: https://doi.org/10.24432/C5XW20}
}

@misc{174969,
  author       = {Wolberg, William and Mangasarian, Olvi and Street, Nick and Street, W.H.},
  title        = {{Breast Cancer Wisconsin (Diagnostic)}},
  year         = {1993},
  howpublished = {UCI Machine Learning Repository},
  note         = {{DOI}: https://doi.org/10.24432/C5DW2B}
}
